\def\cvprPaperID{5929}
\def\confName{ICCV}
\def\confYear{2023}

\def\paperTitle{FateZero: \underline{F}using \underline{A}ttentions for Zero-shot \underline{T}ext-based Video \underline{E}diting}

\def\authorBlock{
Chenyang Qi$^{1}$\thanks{Intern at tencent AI Lab} \qquad 
Xiaodong Cun$^{2}$\footnotemark[2]\qquad
Yong Zhang$^{2}$ \qquad 
Chenyang Lei$^{3}$\\ \qquad 
Xintao Wang$^{2}$ \qquad 
Ying Shan$^{2}$ \qquad 
Qifeng Chen$^{1}$\footnotemark[2]  \qquad \\ \\
$^{1}$HKUST \qquad $^{2}$Tencent AI Lab \qquad $^{3}$CAIR, HKISI-CAS    
\\\\
\url{https://fate-zero-edit.github.io}
}

\newif\ifreview 
\newif\ifarxiv \newcommand{\arxiv}{\arxivtrue}
\newif\ifcamera 
\newif\ifrebuttal

\arxiv % \review OR \arxiv OR \cameraready

\makeatletter
\@namedef{ver@everyshi.sty}{}
\makeatother

\pdfoutput=1
\documentclass[10pt,twocolumn,letterpaper]{article}
\ifreview \usepackage[review]{cvpr} \fi
\ifarxiv \usepackage[pagenumbers]{cvpr} \fi
\ifrebuttal \usepackage[rebuttal]{cvpr} \fi
\ifcamera \usepackage{cvpr} \fi

\usepackage{graphicx}
\usepackage{amsmath}
\usepackage{amssymb}
\usepackage{booktabs}

\usepackage{times}
\usepackage{microtype}
\usepackage{epsfig}
\usepackage[table,xcdraw]{xcolor}
\usepackage{caption}
\usepackage{float}
\usepackage{placeins}
\usepackage{color, colortbl}
\usepackage{stfloats}
\usepackage{enumitem}
\usepackage{tabularx}
\usepackage{xstring}
\usepackage{multirow}
\usepackage{xspace}
\usepackage{url}
\usepackage{subcaption}
\usepackage{xcolor}
\usepackage[hang,flushmargin]{footmisc}

\usepackage{makecell}

\usepackage{algorithm}
\usepackage{algorithmicx}
\usepackage{algpseudocode}
\usepackage{balance}
\usepackage{color}
\usepackage{listings}
\usepackage{array}
\usepackage{bbding}
\usepackage{algpseudocode}
\ifcamera \usepackage[accsupp]{axessibility} \fi

% ...

\definecolor{colorx}{rgb}{0.92,0.49,0.19}

\newcommand{\xiaodong}[1]{{\textcolor{colorx}{[xiaodong: #1]}}}

\newcommand{\chenyang}[1]{\textcolor{blue}{(chenyang: #1)}}

\newcommand{\RM}[1]{}

\ifarxiv  \fi

\newcommand{\R}[1]{{%
    \textbf{%
        \ifstrequal{#1}{1}{\textcolor{red}{R#1}}{%
        \ifstrequal{#1}{2}{\textcolor{blue}{R#1}}{%
        \ifstrequal{#1}{3}{\textcolor{magenta}{R#1}}{%
        \ifstrequal{#1}{4}{\textcolor{teal}{R#1}}{%
                           \textcolor{cyan}{R#1}%
        }}}}%
    }%
}}

\usepackage{tikz}
\newcommand*\circled[1]{\tikz[baseline=(char.base)]{
            \node[shape=circle,draw,inner sep=2pt] (char) {#1};}}  % Add packages to _macros.tex

\usepackage{xr-hyper}

\makeatletter
\newcommand*{\addFileDependency}[1]{
  \typeout{(#1)}
  \@addtofilelist{#1}
  \IfFileExists{#1}{}{\typeout{No file #1.}}
}

\makeatother

\usepackage[pagebackref,breaklinks,colorlinks]{hyperref}
\usepackage[capitalize]{cleveref}
\crefname{section}{Sec.}{Secs.}
\crefname{table}{Table}{Tables}
\crefname{figure}{Fig.}{Figs.}

\definecolor{alizarin}{rgb}{0.82, 0.1, 0.26}

\hypersetup{
    colorlinks = true,
    linkbordercolor = {white},
    citecolor=blue,
    urlcolor=alizarin
}

\begin{document}
%% TITLE
\title{\paperTitle}
\author{\authorBlock}

% \maketitle
\twocolumn[{
\maketitle
\begin{center}
    \captionsetup{type=figure}
    \vspace{-1em}
\newcommand{\imwidth}{1.0\textwidth}

% \setcellgapes{0.5em}
% \makegapedcells
% \vspace{-1em}
\begin{tabular}{@{}c@{}}
  % Prompt Driving Video (top) and Result (bottom) \\
% \parbox{\cellwidth}
% {a silver jeep driving down a snow-covered road in the countryside} &
% {a snow-covered road} &
\parbox{\imwidth}{\includegraphics[width=\imwidth, ]{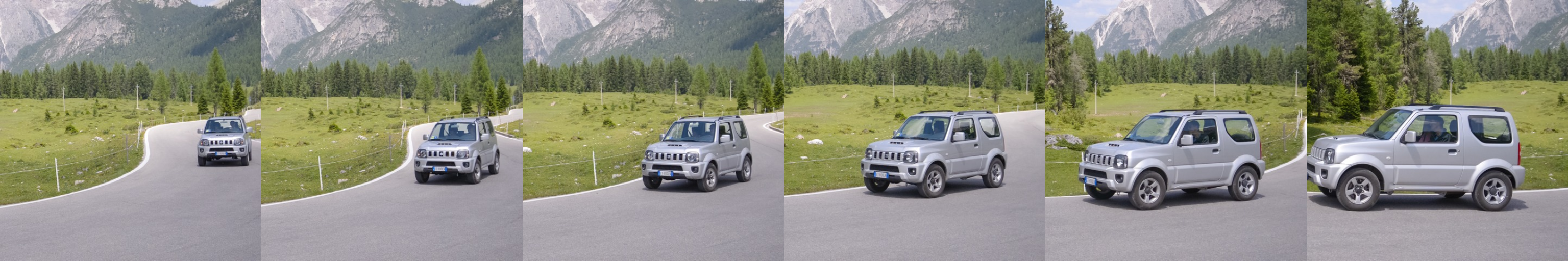}}
\\
% {Source Video Prompt: “A silver jeep driving down a curvy road in the countryside”
% {Source Video Prompt: \texttt{A silver jeep driving down a curvy road in the countryside.}
{Source Video Prompt: {A silver jeep driving down a curvy road in the countryside.}
}
\\
% \parbox{\cellwidth}{car on a snow-covered road in the countryside} &
\parbox{\imwidth}{\includegraphics[width=\imwidth, ]{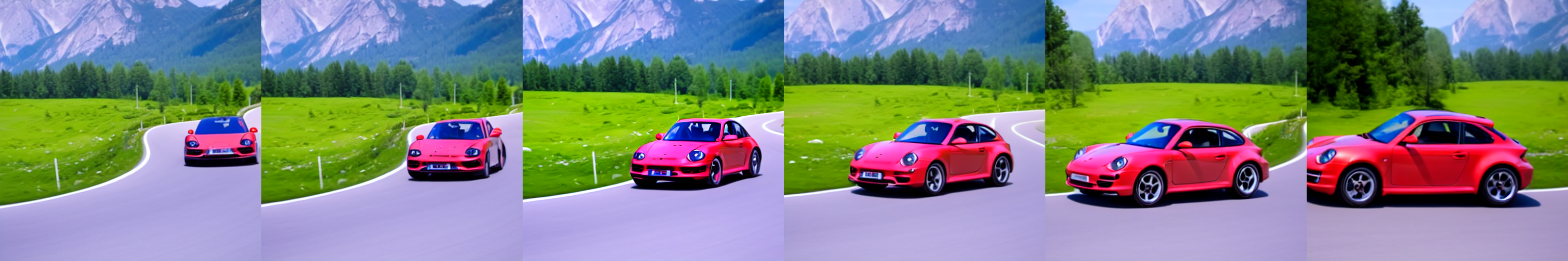}}
\\
{Zero-shot object shape editing with pre-trained video diffusion model~\cite{tuneavideo}: {silver jeep} $\xrightarrow{}$ \textcolor{red}{Porsche car}.
}
\\
\parbox{\imwidth}{\includegraphics[width=\imwidth, ]{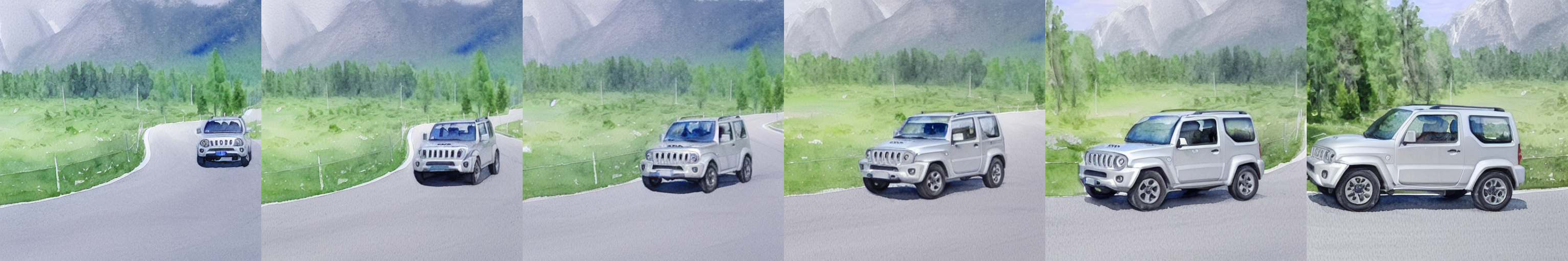}}
\\
{Zero-shot video style editing with pre-trained image diffusion model~\cite{stable-diffusion}: \textcolor{red}{watercolor painting}.
}

\vspace{1em}
\end{tabular}
    \vspace{-0.5em}
    \captionof{figure}{\textbf{Zero-shot text-driven video editing.} We present a zero-shot approach for shape-aware local object editing and video style editing from pre-trained diffusion models~\cite{tuneavideo, stable-diffusion} without any optimization for each target prompt.}
    \label{fig:teaser}
\end{center}
}]

% \maketitle
%%

% {
%   \renewcommand{\thefootnote}%
%     {\fnsymbol{footnote}}
%   \footnotetext[1]{Equal contribution, interns at Microsoft Research.} \footnotetext[2]{Corresponding author.}
% }

\renewcommand{\thefootnote}{\fnsymbol{footnote}}
% \footnotetext[2]{Work done during an internship at Tencent AI Lab.}
\footnotetext[1]{~Work done during an internship at Tencent AI Lab.}
\footnotetext[2]{~Corresponding Authors.}

\begin{abstract}
% \vspace{-1em}
The diffusion-based generative models have achieved remarkable success in text-based image generation. However, since it contains enormous randomness in generation progress, it is still challenging to apply such models for real-world visual content editing, especially in videos. 
In this paper, we propose \texttt{FateZero}, a zero-shot text-based editing method on real-world videos without per-prompt training or use-specific mask. 
\RM{Specifically, different from a pipeline of two independent inversion and then generation stages, we find the intermediate attention maps during inversions store better structure and motion information. We thus reform them to temporally casual attention and replace them in the generation progress. To further reduce the unnecessary semantic leakage of source video and enhance the editing quality, we then remix the temporally casual attentions via the cross-attention features of the source prompt as the mask.}
To edit videos consistently, we propose several techniques based on the pre-trained models. Firstly, in contrast to the straightforward DDIM inversion technique, our approach captures intermediate attention maps during inversion, which effectively retain both structural and motion information. These maps are directly fused in the editing process rather than generated during denoising. To further minimize semantic leakage of the source video, we then fuse self-attentions with a blending mask obtained by cross-attention features from the source prompt. Furthermore, we have implemented a reform of the self-attention mechanism in denoising UNet by introducing spatial-temporal attention to ensure frame consistency.
Yet succinct, our method is the first one to show the ability of zero-shot text-driven video style and local attribute editing from the trained text-to-image model. We also have a better zero-shot shape-aware editing ability based on the text-to-video model~\cite{tuneavideo}. \RM{Besides video, our unified method also achieves state-of-the-art performance in zero-shot image editing.\chenyang{Need exp or remove the zero-shot image}} Extensive experiments demonstrate our superior temporal consistency and editing capability than previous works.
% The code will be released.
% \chenyang{emphasize: our observation at inversion time} \xiaodong{replacing the bold part to the actual pipeline: \textbf{Specifically, we work on replacing and mixing the attention maps between the inversion and generation since the self-attention map keeps the structure of the original natural image and the cross-attention is semantic-related, after remixing, we replace them in the corresponding generation steps for denoising.}}
% \footnote{Since there is no general video diffusion model is publicly available, we use one-shot video generation method~(Tune-A-Video~\cite{tuneavideo}) as the pretrained video diffusion model for zero-shot video editing\xiaodong{can be removed if we actually zero-shot on video}.}.
\end{abstract}

\vspace{-2em}  % ?
\section{Introduction}
\label{sec:intro}
% AIGC is hot and popular, and real image editing is still unclear.
Diffusion-based models~\cite{ddpm} can generate diverse and high-quality images~\cite{imagen, stable-diffusion, dalle2} and videos~\cite{imagen-video, make-a-video, magic-video, lvdm} through text prompts. It also brings large opportunities to edit real-world visual content from these generative priors. 

Previous or concurrent diffusion-based editing methods~\cite{blended,blended_latent,diffedit,pix2pix-zero,pnp,p2p} majorly work on images. To edit real images, their methods utilize deterministic DDIM~\cite{ddim} for the  image-to-noise inversion, and then, the inverted noise gradually generates the edited images under the condition of the target prompt. Based on this pipeline, several methods have been proposed in terms of cross-attention guidance~\cite{pix2pix-zero}, plug-and-play feature~\cite{pnp}, and optimization~\cite{null, imagic}. 
% These methods require use-specific editing mask~\cite{blended,blended_latent} or need additional optimization on the specifically given image~\cite{imagic, null}. 

% Since video needs to keep temporal consistency, these methods may fail in \textit{real} video editing.  \chenyang{Fig. ~comparison. Do exp to support the claim}

% \xiaodong{it is hard for video editing?} 
% Stylized editing (vtoonify, ebsynth) -> large motion inconsistency, specific domain, reference-based
% local editing (layeraltas, text2live)-> optimization needed. altas may failed.
% object edting (shape-aware) -> current method artifacts.
Manipulating videos through generative priors as image editing methods above contains many challenges~(Fig.~\ref{fig:baseline}). First, there are no publicly available generic text-to-video models~\cite{imagen-video,make-a-video}. 
% \yong{compare with dreamix}
Thus, a framework based on image models can be more valuable than on video ones~\cite{dreamix}, thanks to the various open-sourced image models in the community~\cite{t2i-adaptor,controlnet,stable-diffusion,civitai_website}.
However, the text-to-image models~\cite{stable-diffusion} lack the consideration of temporal-aware information, \eg, motion and 3D shape understanding. Directly applying the image editing methods~\cite{sdedit,null} to the video will show obverse flickering. 
Second, although we can use previous video editing methods~\cite{layeraltas,text2live,shape-aware-editing} via keyframe~\cite{ebsynth} or atlas editing~\cite{layeraltas,text2live}, these methods still need atlas learning~\cite{layeraltas,text2live}, keyframe selection~\cite{ebsynth}, and per-prompt tunning~\cite{text2live,shape-aware-editing}. Moreover, while they may work well on the attribute~\cite{layeraltas,text2live} and style~\cite{ebsynth} editing, the shape editing is still a big challenge~\cite{shape-aware-editing}. Finally, as introduced above, current editing methods use DDIM for inversion and then denoising via the new prompt. However, in video inversion, the inverted noise in the $T$ step might break the motion and structure of the original video because of error accumulation (Fig.~\ref{fig: attention mixing} and~\ref{fig:ablation_masked_attention}).

In this paper, we propose \texttt{FateZero}, a simple yet effective method for zero-shot video editing since we do not need to train for each target prompt individually~\cite{text2live,layeraltas,shape-aware-editing} and have no user-specific mask~\cite{blended,blended_latent}.
% allows for direct editing of both the video style and local attributes of real videos by a pretrained text-to-image model~(\eg, Stable diffusion~\cite{stable-diffusion}). Also, given a pretrained video diffusion model~\cite{tuneavideo}, our method can be used in test time for shape editing directly.
% \xd{
Different from image editing, video editing needs to keep the temporal consistency of the edited video, which is not learned by the original trained text-to-image model. We tackle this problem by using two novel designs. Firstly, instead of solely relying on inversion and generation~\cite{p2p,pnp,null}, we adopt a different approach by storing all the self and cross-attention maps at every step of the inversion process. This enables us to subsequently replace them during the denoising steps of the DDIM pipeline. Specifically, we find these self-attention blocks store better motion information and the cross-attention can be used as a threshold mask for self-attention blending spatially. This attention blending operation can keep the original structures unchanged. Furthermore, we reform the self-attention blocks to the spatial-temporal attention blocks as in~\cite{tuneavideo} to make the appearance more consistent.
% }
% In detail, inspired by Tune-A-Video\cite{tuneavideo}, we find self-attention in the stable diffusion's denoising UNet~\cite{unet} storing the motion and structure information, while cross-attention storing the semantic layout. Directly learning~\cite{tuneavideo} or replacing via the DDIM reconstruction feature~\cite{pnp} will still hard to keep the temporal consistency in the generated video.
% Therefore, to address the issue of simple DDIM inversion, 
% Thus, we propose to store all self-attentions and cross-attention at each step of inversion. This enables us to subsequently replace them during the denoising steps of the DDIM pipeline. Besides, we also use the cross-attention maps of the source video prompt as the threshold mask of self-attention remixing, since we want to achieve local editing and keep the original information in other areas unchanged. We perform this attention remixing and fusion in blocks of the UNet, and steps of the inversion and generation. 
% \yong{NO mention about how to solve the temporal consistency.}
Powered by our novel designs, we can directly edit the style and the attribute of the real-world video (Fig.~\ref{fig:exp_attribute_style_edit}) using the pre-trained text-to-image model~\cite{stable-diffusion}. Also, after getting the video diffusion model~(\eg, pretrained Tune-A-Video~\cite{tuneavideo}), our method  shows better object editing (Fig.~\ref{fig:exp_swan_shape_edit}) ability in test-time than simple DDIM inversion~\cite{ddim}. The extensive experiments provide evidence of the advantages offered by the proposed method for both video and image editing. 
% \xiaodong{can be moved if we do not contain image samples.}

Our contributions are summarized as follows:
% We summarize the contribution of our methods as follows:
% \xiaodong{contribution need to be polished.}
\begin{itemize}
    \item We present the first framework for temporal-consistent zero-shot text-based video editing using pretrained text-to-image model.
    % \yong{check `the first'?}
    % \item We propose a new editing method via remixing the self-attention maps in the inversion process and generation process using the source prompt's cross-attention map. \yong{emphasize issues: semantic leakage and consistency }
    \item We propose to fuse the attention maps in the inversion process and generation process to preserve the motion and structure consistency during editing.
    \item Our novel Attention Blending Block utilizes the source prompt's cross-attention map during attention fusion to prevent source semantic leakage and improve the shape-editing capability.
    \item We show extensive applications of our method in video style editing, video local editing, video object replacement, \etc.
    % \chenyang{if no exp, downclaim}
    % Experiments show that our framework outperforms state-of-the-art editing methods with better editing quality, temporal consistency and image fidelity.
\end{itemize}

\section{Related Work}
\label{sec:related}

\begin{figure*}[h]
    \centering
    \includegraphics[width=\textwidth]{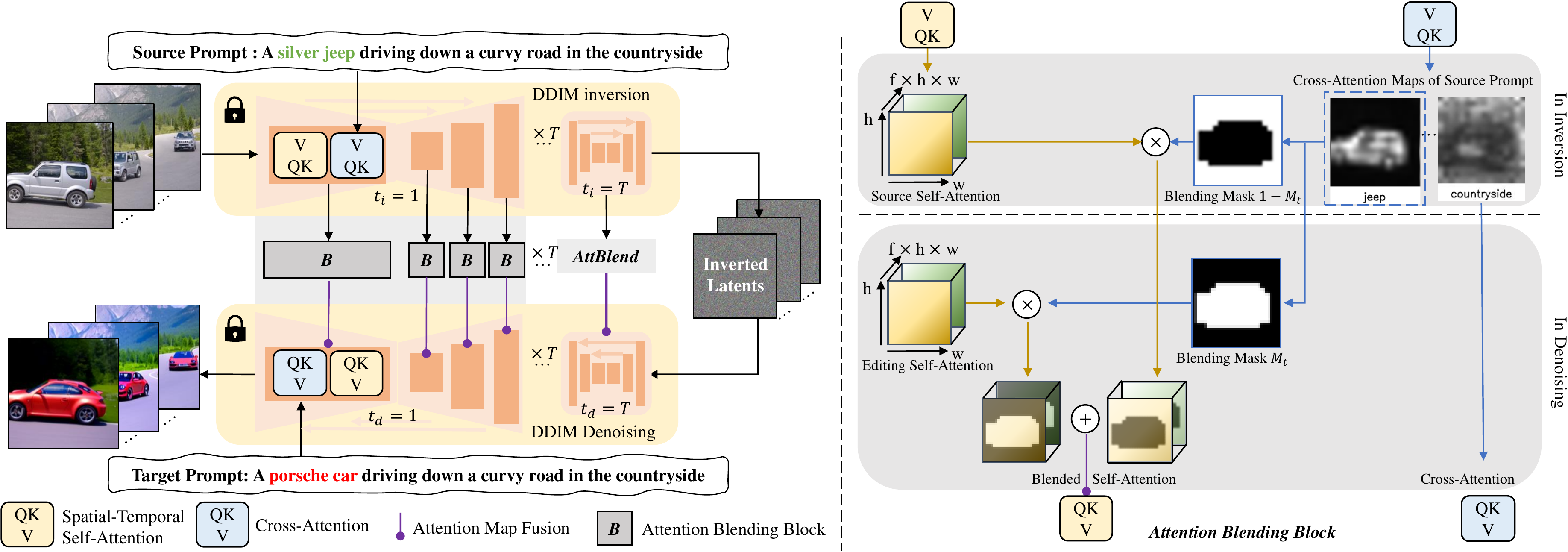}
    % \vspace{0.5em}
    \caption{\textbf{The overview of our approach}.
    % Given an input source video $x=\{x^0,x^1, ... x^c\}$ with clip length $c$, we first encode the clip to its latent representation $z=\{z^0,z^1, ... z^c\}$ with the encoder $\mathcal{E}$ in frame-wise manner. 
    % The input is source prompt p_src
    Our input is the user-provided source prompt $p_{src}$,  target prompt $p_{edit}$ and clean latent $z=\{z^1,z^2, ... z^n\}$ encoded from input source video $x=\{x^1,x^2, ... x^n\}$ with number frames $n$ in a video sequence.
    On the \textbf{left}, we first invert the video using DDIM inversion pipeline into noisy latent $z_T$ using the source prompt $p_{src}$ and an inflated 3D U-Net $\varepsilon_\theta$. During each inversion timestep $t$, we store both spatial-temporal self-attention maps $s^{src}_t$ and cross-attention maps $c^{src}_t$.
   At the editing stage of the DDIM denoising, we denoise the latent $z_T$ back to clean image $\hat{z}_0$ conditioned on target prompt $p_{edit}$. At each denoising timestep $t$ , we fuse the attention maps ($s^{edit}_t$ and $c^{edit}_t$) in $\varepsilon_\theta$ with stored attention map ($s^{src}_t$, $c^{src}_t$) using the proposed Attention Blending Block. \textbf{Right}: Specifically, we replace the cross-attention maps $c^{edit}_t$ of un-edited words~(\eg, road and countryside) with source maps $c^{src}_t$ of them. In addition, we blend the self-attention map during inversion $s^{src}_t$ and editing $s^{edit}_t$ with an adaptive spatial mask obtained from cross-attention $c^{src}_t$, which represents the areas that the user wants to edit. 
    % transform $y$ to DCT coefficients $C$ and predict the quantization tables $Q_L, Q_C$ with our quantization prediction module (QPM). Third, we adopt an entropy model~\cite{balle2017end} to estimate the bitrate of the quantized coefficients $\widetilde{C}$ at training stage. After the rounding and truncation, which we denoted as $[ \cdot ]$, the $[Q_L], [Q_C]$ and $[\widetilde{C}]$ can be written and read with off-the-shelf JPEG API at the testing stage. To restore the HR, we extract features from $\widehat{C}$ with a frequency feature extractor $f$ and produce the high-fidelity image $\hat{x}$ with the decoder $D$.}
    }
    % \vspace{-1em}
    \label{fig:main_framework}
\end{figure*}

\noindent\textbf{Video Editing.}
% Previous video editing works aim to change the appearance of the given images. \eg,
% Ebsynth~\cite{ebsynth} synthesis coherence video contents via the patch-based convolution network and the selected key-frames, however, the editing is only limited to the ranges of the images. VideoToonify~\cite{Vtoonify},
% Text2Live~\cite{text2live}, layer 
% altas~\cite{layeraltas}.
% Tune-A-Video~\cite{tuneavideo}, gen1~\cite{gen1}, Dreamix~\cite{dreamix}. 
Video can be edited via several aspects. For video stylizing editing, current methods~\cite{stylit, ebsynth} rely on the example as the style guide and these methods may fail when the track is lost. By processing frames individually using image style transfer~\cite{gatys2016image, johnson2016perceptual}, some works also learn to reduce the temporal consistency~\cite{dvp, BTSSPP15, lai2018learning,lei2022deep} in a post-process way. However, the style may still be imperfect since the style transfer only measures the perceptual distance~\cite{zhang2018perceptual}. Several works also show better consistency but on the specific domain, \eg, portrait video~\cite{fivser2017example, Vtoonify}. 
For video local editing, layer-atlas based methods~\cite{layeraltas, text2live} show a promising direction by editing the video on a flattened texture map. 
% However, the 2d atlas is not meaningful since it lacks 3d motion perception, and prompt-specific optimization is required. 
However, the 2d atlas lacks 3d motion perception to support shape editing, and prompt-specific optimization is required. 

A more challenging topic is to edit the object shape in the real-world video. Current method shows obvious artifacts even with the optimization on generative priors~\cite{shape-aware-editing}. The stronger prior of the diffusion-based model also draws the attention of current researchers. \eg, gen1~\cite{gen1} trains a conditional model for depth and text-guided video generation, which can edit the appearance of the generated images on the fly. Dreamix~\cite{dreamix} finetunes a stronger diffusion-based video model~\cite{imagen-video} for editing with stronger generative priors. Both of these methods need privacy and powerful video diffusion models for editing. Thus, the applications of the current larger-scale fine-tuned text-to-image models~\cite{civitai_website} cannot be used directly. 
% Tune-A-Video~\cite{tuneavideo} proposes a one-shot video generation method for single prompt video generation. This method can also generate edited content from audio. However, the generated frames are still not continuous and the ability of real-world video editing is restricted by simple DDIM inversion~\cite{ddim}.
% \chenyang{`the information preserving of source structure is limited by only using latents from DDIM inversion' can be better? }

\noindent\textbf{Image and Video Generation Models.} Image generation is a basic and hot topic in computer vision. Early works mainly use VAE~\cite{vae} or GAN~\cite{gan} to model the distribution on the specific domain.
Recent works adopt VQVAE~\cite{vqvae} and transformer~\cite{taming} for image generation. However, due to the difficulties in training these models, they only work well on the specific domain, \eg, face~\cite{stylegan}. 
On the other hand, the editing ability of these models is relatively weak since the feature space of GAN is high-level, and the quantified tokens can not be considered individually. 
Another type of method focuses on text-to-image generation. DALL-E~\cite{ramesh2021zero, dalle2} and CogView~\cite{ding2021cogview} train an image generative pre-training transformer~(GPT) to generate images from a CLIP~\cite{CLIP1} text embedding. 
Recent models~\cite{stable-diffusion, imagen} benefit from the stability of training diffusion-based model~\cite{ddpm}. These models can be scaled by a huge dataset and show surprisingly good results on text-to-image generation by integrating large language model conditions since its latent space has spatial structure, which provides a stronger edit ability than previous GAN~\cite{stylegan} based methods. 
Generating videos is much more difficult than images. Current methods rely on the larger cascaded models~\cite{imagen-video, make-a-video} and dataset. Differently, magic-video~\cite{magic-video} and gen1~\cite{gen1} initialize the model from text-to-image~\cite{stable-diffusion} and generate the continuous contents through extra time-aware layers.
Recently, Tune-A-Video~\cite{tuneavideo} over-fits a single video for text-based video generation.
After training, the model can generate related motion from similar prompts. However, how to edit real-world content using this model is still unclear. Inspired by the image editing methods and tune-a-video, our method can edit the style of the real-world video and images using the trained text-to-image model~\cite{stable-diffusion} and shows better object replacing performance than the one-shot finetuned video diffusion model~\cite{tuneavideo} with simple DDIM inversion~\cite{ddim} in real videos (Fig.~\ref{fig:baseline}).

\noindent\textbf{Image Editing in Diffusion Model.}
Many recent works adopt the trained diffusion model for editing. 
SDEdit~\cite{sdedit} generates content for a new prompt by adding noise to the image first.
DiffEdit~\cite{diffedit} computes the edit mask by the noise differences of the text prompts, and then, blends the inversion noises into the image generation process. Similar work has also been proposed by Blended Diffusion~\cite{blended,blended_latent}, which combines the features of each step for image blending. Plug-and-play~\cite{pnp} gets the inversion noise and applies the denoising for feature reconstruction. After that, the self-attention features in editing are replaced with that in reconstruction directly. Pix2pix-Zero~\cite{pix2pix-zero} edits the image with the cross-attention guidance. Prompt-to-Prompt~\cite{p2p} proves that images can be edited via reweighting the cross-attention map of different prompts. 
% Not so much related
% InstructPix2Pix~\cite{instructpix2pix} uses the prompt-to-prompt~\cite{p2p} to synthesize large-scale paired datasets. And then, editing the input image via text prompt directly. 
There are also some methods to achieve better editing ability via optimization~\cite{null,imagic}. However, a naive frame-wise application of these image methods to video results in flickering and inconsistency among 
% different 
frames.
% Different from above method, o
% \input{figs/framework}
% \newpage
% page 3.5, end of related work
% \newpage
\RM{\begin{figure}[th]
    \centering
    \includegraphics[width=0.5\textwidth]{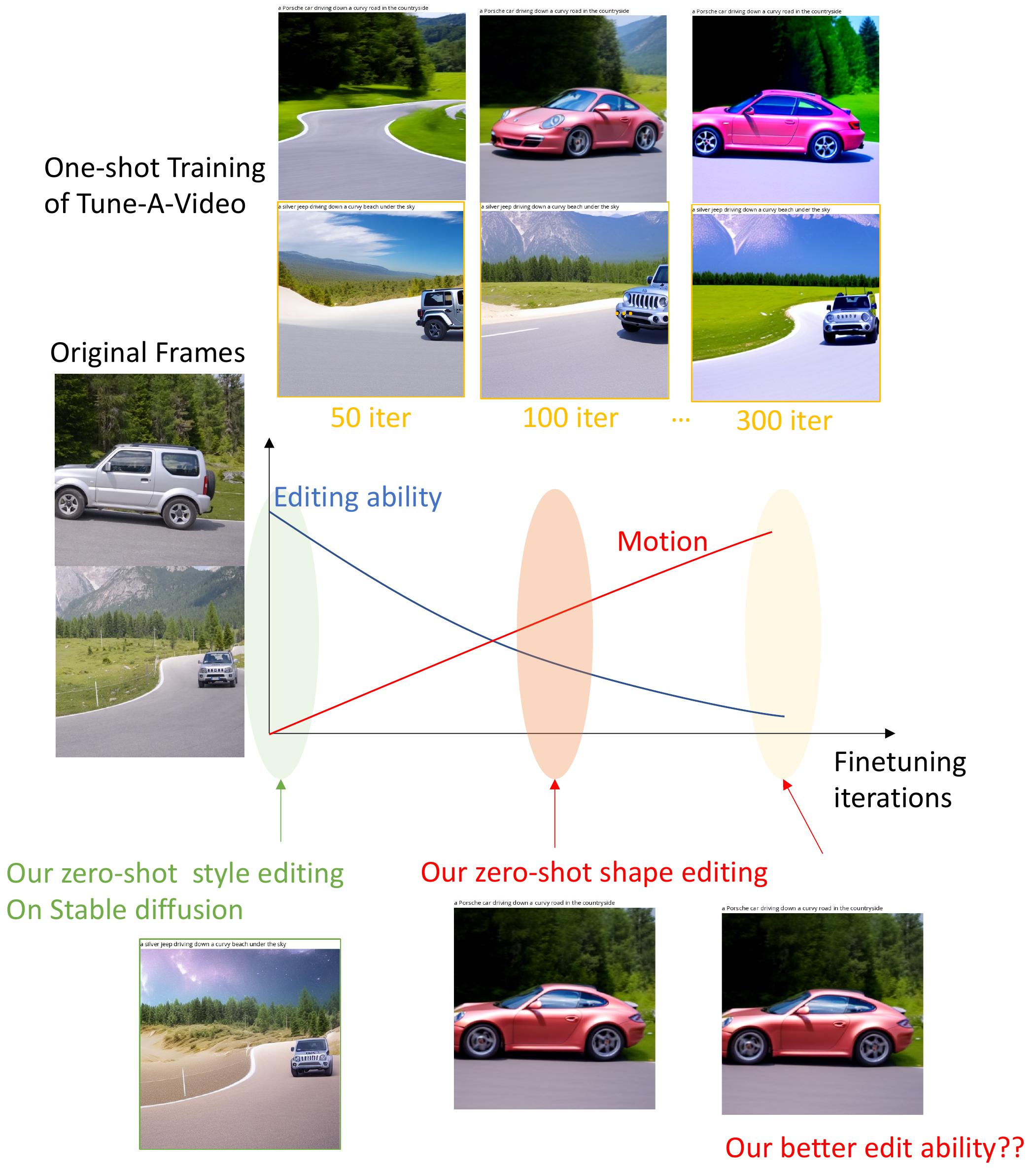}
    \caption{\chenyang{ Placeholder: A tuning stage typically increase the faithfulness to input sequence and motion similarity, decreases the editing ability} \chenyang{inverted attention? inversion time attention?} \xiaodong{if this figure is too hard to represent, we can move it to supp or delete it.}}
    \label{fig:tuning progress}
\end{figure}}
\section{Methods}
\label{sec:method}
% We focus on xxx
% Define the task of style, foreground and background 

% How to achieve both precise localization and diverse editing effects? Different from existing optimization method (tune a video, text2live, shape-ware) zero-shot and short tuning
% (phylosophy of zero-shot editing, why it works?)

\RM{we need to define the zero-shot means we can add/replace words at test time without any training.}

\RM{It is worth mentioning why our method can be used for video editing, specifically due to its temporal consistency and the failure of other methods in achieving this.}

% our setting, from pretrained zero shot no need tuning for style and foreground. Tuning at object -level shape editing and large motion synthesis

% Define the setting of p2p2, notation. provide source, provide target

% Section overview and framework figures
We target zero-shot text-driven video editing~(\eg, style, attribute, and shape) without optimization for each target prompt or the user-provided mask.
% \RM{
% To edit a video, existing methods typically employ an optimization for editing prompt~\cite{shape-aware-editing, text2live} or source video~\cite{tuneavideo}, which is inconvenient and tedious in real applications. Besides, tuning a large model on a short video brings an additional trade-off between motion similarity and editing ability, where the model suffers from catastrophic forgetting after overfitting as in Fig.~\ref{fig:tuning progress}. 
% }
% We find that simple text-based appearance editing, \eg, shape and local attribute editing, can be handled directly via the pretrained text-to-image model~\cite{stable-diffusion} by our new editing method. As for the challenge editing, \eg, shape-aware editing, our method can also gain better performance on the pre-trained one-shot video diffusion model~\cite{tuneavideo} than simple DDIM inversion~\cite{ddim}. 
In Sec.~\ref{sec:invertibility_and_attention}, we first give the details of the latent diffusion and DDIM inversion. After that, we introduce our method that enables video appearance editing (Sec.~\ref{sec:fate-zero}) via the pre-trained text-to-image models~\cite{stable-diffusion}. Finally, we discuss a more challenging case that also enables the shape-aware editing of video using the video diffusion model in Sec.~\ref{sec:shape-aware-editing}. 
Notice that, the proposed method is a general editing method and can be used in various text-to-image or text-to-video models. In this paper, we majorly use Stable Diffusion~\cite{stable-diffusion} and the video generation model based on Stable Diffusion~(Tune-A-Video~\cite{tuneavideo}) for its popularity and generalization ability.

\subsection{Preliminary: Latent Diffusion and Inversion}
\label{sec:invertibility_and_attention}

\noindent\textbf{Latent Diffusion Models~\cite{stable-diffusion}} are introduced to diffuse and denoise the latent space of an autoencoder. First, an encoder $\mathcal{E}$ compresses a RGB image $x$ to a low-resolution latent $z=\mathcal{E}(x)$
% , such that the latent 
, which can be reconstructed back to image $ \mathcal{D}(z) \approx x $ by decoder $\mathcal{D}$.
% by decoder $\mathcal{D}$ to image $\mathcal{D}(z) \approx x$. 
Second, a U-Net~\cite{unet} $\varepsilon_\theta$ 
% composed of 
containing cross-attention and self-attention~\cite{attention_is_all_you_need} is trained to remove the artificial noise using the objective:
\vspace{-0.5em}
\begin{equation}
\min _\theta E_{z_0, \varepsilon \sim N(0, I), t \sim \text { Uniform }(1, T)}\left\|\varepsilon-\varepsilon_\theta\left(z_t, t, p \right)\right\|_2^2,
\end{equation}
% \vspace{-0.2em}
where $p$ is the embedding of the conditional text prompt and $z_t$ is a noisy sample of $z_0$ at timestep $t$.
% with the noise of randomly sampled 

\noindent\textbf{DDIM Inversion~\cite{ddim}.} During inference, deterministic DDIM sampling is employed to convert a random noise $z_T$ to a clean latent $z_0$ in a sequence of timestep $t: T\rightarrow 1$: 
\vspace{-0.5em}
\begin{equation}
\label{eq: denoise}
    z_{t-1} = \sqrt{\alpha_{t-1}} \; \frac{z_t - \sqrt{1-\alpha_t}{\varepsilon_\theta}}{\sqrt{\alpha_t}}+ \sqrt{1-\alpha_{t-1}}{\varepsilon_\theta},
\end{equation}
where $\alpha_{t}$ is a parameter for noise scheduling~\cite{ddim, ddpm}
% \yong{$\alpha$ is computed using $\beta$. $\beta$ controls the scheduling ?}.

\begin{figure}[t]
    \centering
    \includegraphics[width=\columnwidth]{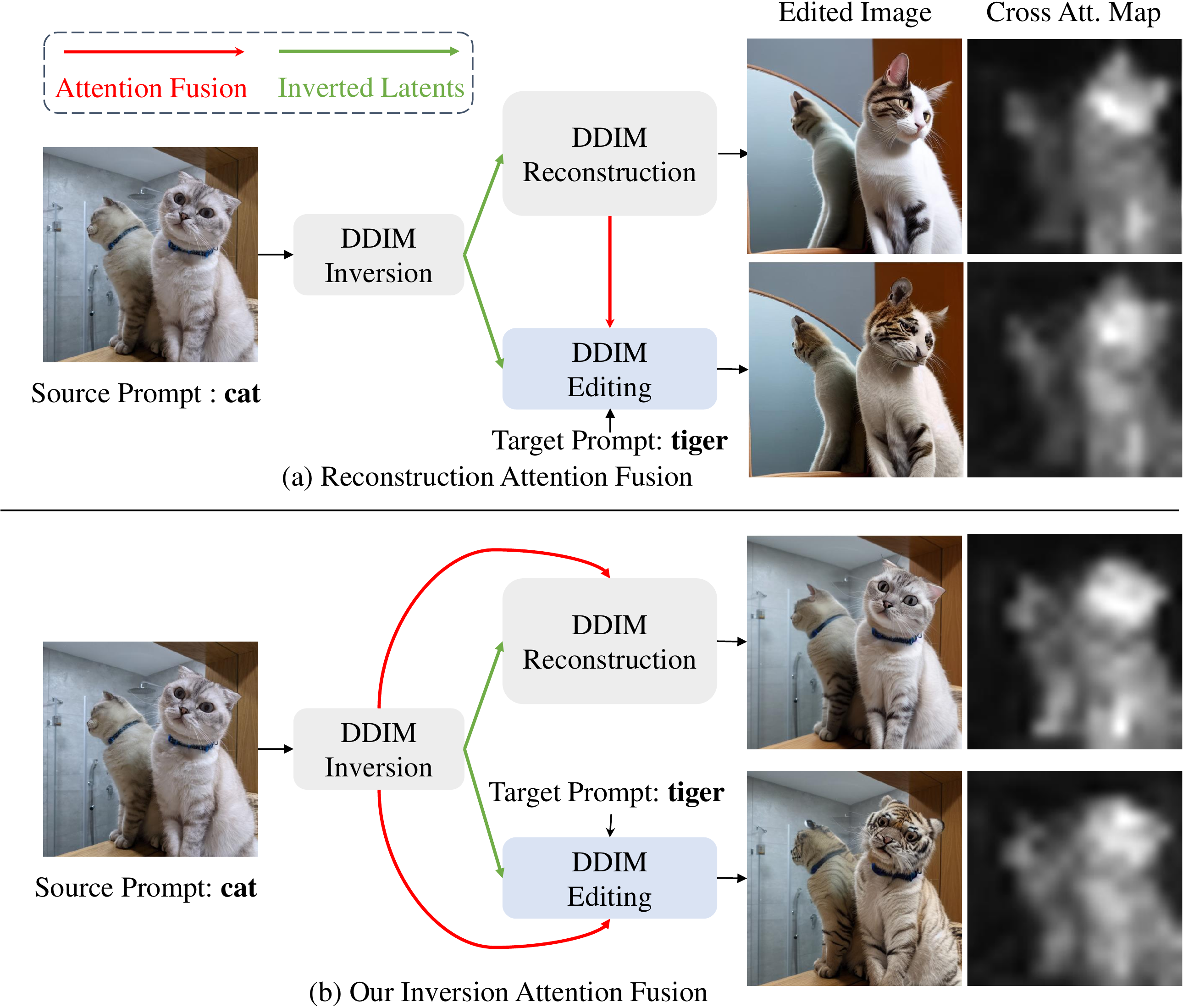}
    \caption{\textbf{Zero-shot local attributed editing (cat $\rightarrow$ \textcolor{red}{tiger}) using stable diffusion.} In contrast to fusion with attention during reconstruction (a) in previous work~\cite{p2p,pnp,pix2pix-zero},
    our inversion attention fusion (b) provides more accurate structure guidance and editing ability, as visualized on the right side.
    }
    % \vspace{-1em}
    % \xiaodong{add a,b to the two figure} 
    \label{fig:attention comparison}
\end{figure}

Based on the ODE limit analysis of the diffusion process, DDIM inversion~\cite{ddim,DiffusionBeatGANs} is proposed to map a clean latent $z_0$ back to a noised latent $\hat{z}_T$ in revered steps  $t: 1\rightarrow T$:
\begin{equation}
\label{eq: add noise}
    \hat{z}_{t} = \sqrt{\alpha_{t}} \; \frac{\hat{z}_{t-1} - \sqrt{1-\alpha_{t-1}}{\varepsilon_\theta}}{\sqrt{\alpha_{t-1}}} + \sqrt{1-\alpha_{t}}{\varepsilon_\theta}.
\end{equation}

Such that the inverted latent $\hat{z}_T$ can reconstruct a latent $\hat{z}_0(p_{src}) = \text{DDIM}(\hat{z}_T, p_{src}) $ similar to the clean latent $z_0$ at classifier-free guidance scale $s_{cfg} = 1$. Recently, image editing methods~\cite{p2p,null,pnp,pix2pix-zero} use a large classifier-free guidance scale $s_{cfg} \gg 1$ to edit the latent as $\hat{z}_0(p_{edit}) = \text{DDIM}(\hat{z}_T, p_{edit})$ (second row in Fig~\ref{fig:attention comparison}(a)), where a reconstruction of $\hat{z}_0(p_{src})$ is conducted in parallel to provide attention constraints. (first row in Fig~\ref{fig:attention comparison}(a)).

% \yong{Depict the reconstruction process.}

% \xiaodong{need to specific to that we are working on video.}

\RM{
However, as shown in the DDIM reconstruction of Fig.~\ref{fig:attention comparison}~(a), a naive DDIM inversion and reconstruction with $s_{cfg} \gg 1$ lose the layout (mirrors) and inner structure (cat) of objects in the image.

To constrain the structure of the generated image, Prompt2Prompt~\cite{p2p}
% replaces the cross-attention maps of U-Net $\varepsilon_\theta$ during the editing with the maps stored during the reconstructed time. 
edits the cross-attention maps of U-Net $\varepsilon_\theta$ during editing. 

More Formally, for each cross-attention block, the attention maps of the input spatial features $\phi(z_t)$ are
% are projected into a query matrix $Q_c(\phi(z_t)$ and key $K_c(p_{edit})$ matrics
% the attention maps are
\begin{equation}
    M_c=\operatorname{Softmax}\left(\frac{Q_c(\phi(z_t)) K_c(p)^T}{\sqrt{d}}\right),
\end{equation}
where $Q_c(\phi(z_t))$ is the query matrix of spatial features, $K_c(p_{edit})$ is the key matrix of prompt $p$. During editing, Replacing the maps $M_c(p_{edit})$ with those from reconstruction $M_c(p_{src})$ results in an edited image, which has a similar semantic layout as the reconstructed one, as shown in the Fig.~\ref{fig:attention comparison}~(a). 
}

% \subsection{FateZero Editing on Styles and Local Attributes}
\subsection{FateZero Video Editing}
\label{sec:fate-zero}
As shown in Fig.~\ref{fig:main_framework}, we 
% also 
use the pretrained text-to-image model, \ie, Stable Diffusion, as our base model, which contains a UNet for $T$-timestep denoising.
% a $T$ timestamps UNet for denoising.
Instead of straightforwardly exploiting the regular pipeline of latent editing guided by reconstruction attention, we have made several critical modifications for video editing as follows.
% Although we typically adhere to the pipeline of inversion and editing, 
% we have made several modifications as listed below.

\noindent\textbf{Inversion Attention Fusion.}
Direct editing using the inverted noise results in frame inconsistency, which may be attributed to two factors. First, the invertible property of DDIM discussed in Eq.~\eqref{eq: denoise} and Eq.~\eqref{eq: add noise} only holds in the limit of small steps~\cite{ddim,ncsn}.
% ~\xiaodong{any evidence or citation?}
Nevertheless, the present requirements of 50 DDIM denoising steps lead to an accumulation of errors with each subsequent step. 
Second, 
% a large classifier-free guidance $s_{cfg} \gg 1$ will increase the edit ability in denoising, which also brings inconsistency between inversion and editing, while a small $s_{cfg} = 1$ does not have enough editing ability. 
using a large classifier-free guidance $s_{cfg} \gg 1$ can increase the edit ability in denoising, but the large editing freedom leads to inconsistent neighboring frames.
Therefore, previous methods require optimization of text-embedding~\cite{p2p} or other regularization~\cite{pix2pix-zero}.  

% , where only a single frame is involved, 
While the issues seem trivial in the context of single-frame editing
they can become magnified when working with video as even minor discrepancies among frames will be accentuated along the temporal indexes.
% To solve this problem, we begin with the analysis of attention maps in the denoising diffusion model, which is typically used in previous image editing methods~\cite{p2p,pnp,pix2pix-zero}.

% Although the attention maps during reconstruction have been utilized~\cite{p2p, pnp, pix2pix-zero} to control the semantic layout of the generated images, an efficient and high-quality source image reconstruction without any optimization is still a challenge. 
% We find. First, the invertible property of DDIM discussed in Eq.~\eqref{eq: denoise} and Eq.~\eqref{eq: add noise} only holds in the limit of small steps. However, current methods set the step as 50 to accelerate the pipeline, which brings an accumulation of errors in each time step. Secondly, a large classifier-free guidance $s_{cfg} \gg 1$ also brings inconsistency between inversion and editing, while a small $s_{cfg} = 1$ does not have enough editing ability. Therefore, previous methods require optimization of text-embedding ~\cite{p2p} or other regularization~\cite{pix2pix-zero}.

% In contrast with previous methods, 
To alleviate these issues, our framework utilizes the attention maps during inversion steps~(Eq.~\eqref{eq: add noise}), which is available because the source prompt $p_{src}$
% $p_{src}$ 
and initial latent $z_0$
% $\mathcal{E}(x) = z_0$ 
are provided to the UNet 
% $\varepsilon_\theta$ 
during inversion. 
% As observed in Fig.~\ref{fig:attention comparison}~(b), the cross-attention map during inversion captures the silhouette and the pose of the cat in the source image, while the map during reconstruction has a noticeable difference. The gap between inversion and reconstruction is more sensitive in zero-shot video editing as shown in the Fig.~\ref{fig:ablation_video_inversion_attn}.
% , since there is yet no video diffusion model available for high-quality inversion.
Formally, during inversion, we store the intermediate self-attention maps $[s_{t}^{\text{src}}]_{t=1}^T$, cross-attention maps $[c_{t}^{\text{src}}]_{t=1}^T$ at each timestep $t$ and the final latent feature maps $z_T$ as
\begin{equation}
\vspace{-0.5em}
z_T, [c_{t}^{\text{src}}]_{t=1}^T, [s_{t}^{\text{src}}]_{t=1}^T=\Call{DDIM-Inv}{z_0, p_{src}},
% \vspace{-0.5em}
\end{equation}
where $\Call{DDIM-Inv}{}$ stands for the DDIM inversion pipeline discussed in Eq.~\eqref{eq: add noise}.
% \yong{no description of DDIMInv}
During the editing stage, 
% we predict the noise to remove by fusion 
we can obtain the noise to remove by fusing
the attention from inversion: 
\begin{equation}
\hat{\epsilon}_t = \Call{Att-Fusion}{\varepsilon_\theta, z_t, t, p_{\text{edit}},c_{t}^{\text{src}}, s_{t}^{\text{src}}}.
\end{equation}
where $p_\text{edit}$ represents the modified prompt. In function $\Call{Att-Fusion}{}$, we inject the cross-attention maps of the unchanged part of the prompt similar to Prompt-to-Prompt~\cite{p2p}. We also replace self-attention maps to preserve the original structure and motion during the style and attribute editing.
% \yong{No words to depict how to perform att-fusion.}
\RM{where latents $z_T$ has a shape of $n\times c \times h \times w  = 8\times4\times64\times64$ in our implementation. 
Cross attention maps $c_t^{src}$ has the shape of  
$ \text{frames}  \times \text{attention\_head} \times \text{pixels} \times \text{text\_token}  = 8\times8\times\{hw\}\times77$, which is correspondence between each pixel and each word token.
Similarly, self-attention maps have the shape of  $ \text{frames}  \times \text{attention\_head} \times \text{pixels} \times \text{pixels} \}  = 8\times8\times\{hw\}\times\{hw\}$, which represent the inner structure in each frame and implicit warping between frames at different temporal index.}

Fig.~\ref{fig:attention comparison} shows a toy comparison example between our attention fusion method and the typical method with simply inversion and then generation as in \cite{p2p, null} for image editing.
% In Fig.~\ref{fig:attention comparison}~(b), we replace the cross-attention and self-attention maps following prompt-to-prompt~\cite{p2p}.
The cross-attention map during inversion captures the silhouette and the pose of the cat in the source image, but the map during reconstruction has a noticeable difference. While in the video, the attention consistency might influence the temporal consistency as shown in Fig.~\ref{fig:ablation_inversion_attn}. This is because the spatial-temporal self-attention maps represent the correspondence between frames and the temporal modeling ability of existing video diffusion model~\cite{tuneavideo} is not satisfactory.
% \yong{Depict the key method difference between the two.
% }
% Compared with the result from reconstruction attention injection and optimization-based image editing method~\cite{null}, our inversion attention injection builds a strong zero-shot editing method.
% ablation study show that xxx
% ~\ref{ablation:}

% \newpage

\begin{figure}[t]
    \centering
    \includegraphics[width=0.47\textwidth]{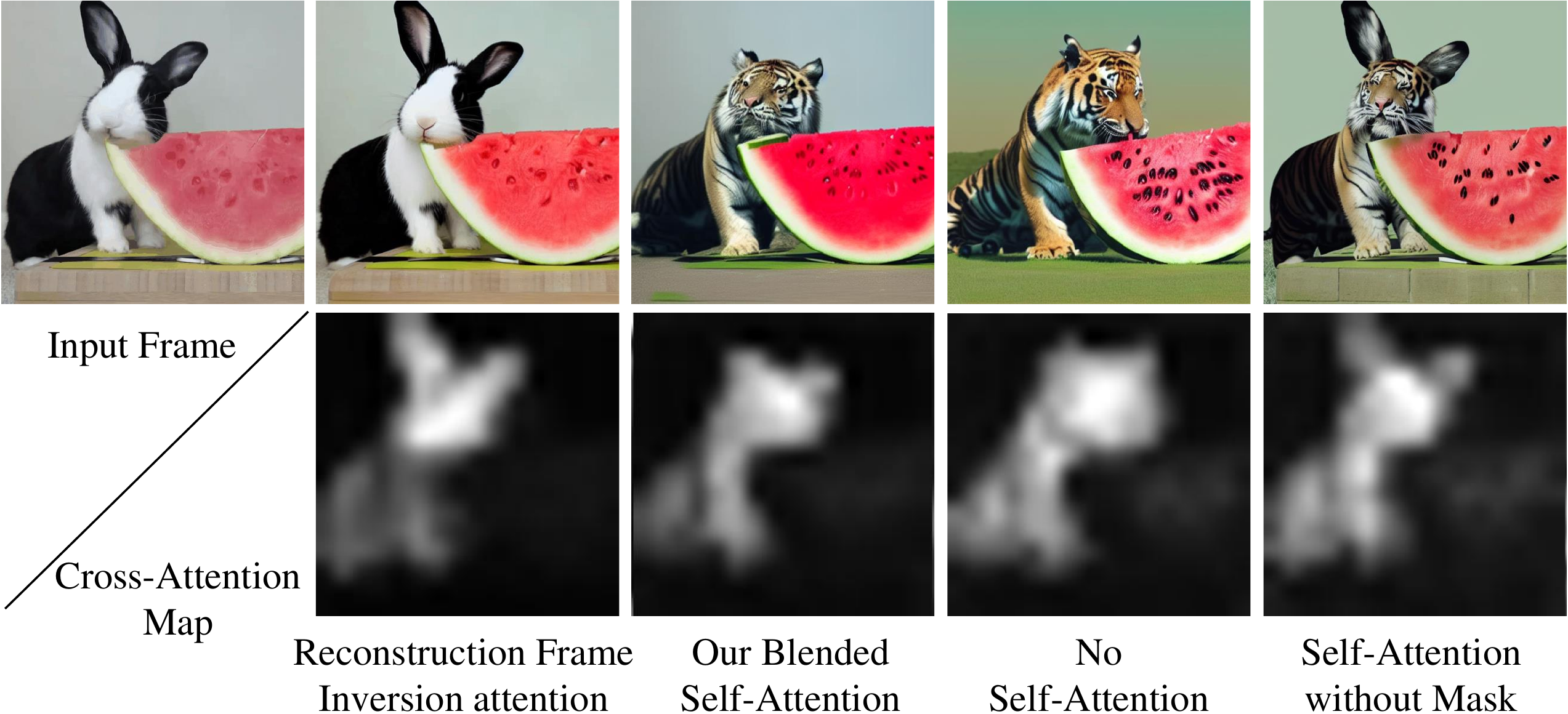}
    \caption{\textbf{Study of blended self-attention in zero-shot shape editing (rabbit $\rightarrow$ \textcolor{red}{tiger}) using stable diffusion}. Forth and fifth columns: Ignoring self-attention can not preserve the original structure and background, and naive replacement causes artifacts. Third column: Blending the self-attention using the cross-attention map (the second row) obtains both new shape from the target text with a similar pose and background from the input frame. 
    }
    % \vspace{-1em}
    % \xiaodong{check all the caption}}
    \label{fig: attention mixing}
\end{figure}

\noindent\textbf{Attention Map Blending.}
\label{sec:Attention map mixing}
% Existing attention fusion methods~\cite{p2p,null} replace full-resolution cross-attention maps ($(hw) \times (token_num)$) and self-attention maps $(hw) \times (hw)$ to preserve the background and object structure in the source image, which is presented in Fig.~\ref{fig:attention comparison}.
% shows that this full-resolution fusion edits the attribute without changing the pose structure.
Inversion-time attention fusion might be insufficient in local attrition editing, as shown in an image example in Fig.~\ref{fig: attention mixing}. In the third column, replacing self-attention $s^{edit} \in \mathbb{R}^{hw \times hw} $ with $s^{src}$ brings unnecessary structure leakage and the generated image has unpleasant blending artifacts in the visualization. On the other hand, if we keep $s^{edit}$ during the DDIM denoising pipeline, the structure of the background and watermelon has unwanted changes, and the pose of the original rabbit is also lost. Inspired by the fact that the cross-attention map provides the semantic layout of the image~\cite{p2p}, as visualized in the second row of Fig.~\ref{fig: attention mixing}, we obtain a binary mask $M_t$ by thresholding the cross-attention map of the edited words during inversion by a constant $\tau$~\cite{blended,blended_latent}. Then, the self-attention maps of editing stage $s^{edit}_t$ and inversion stage $s^{src}_t$ are blended with the binary mask $M_t$, as illustrated in Fig.~\ref{fig:main_framework}. Formally, the attention map fusion is implemented as
% \begin{align}
% M_t = \Call{HeavisideStep}{c_t^{src},  t},\\
% s_{t}^{\text{fused}} = M_t \odot s_{t}^{\text{tgt}} + (1 - M_t) \odot s_{t}^{\text{src}}
% \end{align}
\begin{eqnarray}
\vspace{-2em}
M_t &=& \Call{HeavisideStep}{c_t^{src},  \tau},\\
s_{t}^{\text{fused}} &=& M_t \odot s_{t}^{\text{edit}} + (1 - M_t) \odot s_{t}^{\text{src}}.
\vspace{-2em}
\end{eqnarray}
% where $M_t$ is a binary mask obtained by threshold parameter $\tau$.

\noindent\textbf{Spatial-Temporal Self-Attention.}
% \xiaodong{may more citation}
The previous two designs make our method a strong editing method that can preserve the better structure, and also a big potential in video editing. However, denoising each frame individually still produces inconsistent video. Inspired by the casual self-attention~\cite{attention_is_all_you_need,lvdm,vdm,Phenaki} and recent one-shot video generation method~\cite{tuneavideo}, we reshape the original self-attention to Spatial-Temporal Self-Attention without changing pretrained weights. Specifically, we implement $\Call{Attention}{Q, K, V}$ for feature $z^i$ at temporal index $i \in [1, n]$ as
\vspace{-0.5em}
\begin{equation}
    % Q=W^Q \mathbf{z}^i, K=W^K\left[\mathbf{z}^{i}; \mathbf{z}^{\Call{w}{i}}\right], V=W^V\left[\mathbf{z}^{i}; \mathbf{z}^{\Call{w}{i}}\right]
    Q=W^Q \mathbf{z}^i, K=W^K\left[\mathbf{z}^{i}; \mathbf{z}^{\text{w}}\right], V=W^V\left[\mathbf{z}^{i}; \mathbf{z}^{\text{w}}\right],
\end{equation}
where $[\cdot]$ denotes the concatenation operation and $W^Q$, $W^K$, $W^V$ are the projection matrices from pretrained model. 
% $\Call{w}{i}$ is the temporal index to be warped for the current temporal index $i$. 
% $\Call{w}{i}$ is the temporal index to be warped for the current temporal index $i$. 
Empirically, we find it is enough to warp the middle frame $\mathbf{z}^{\text{w}} = z^{\text{Round}[\frac{n}{2}]}$ for attribute and style editing. Thus, the spatial-temporal self-attention map is represented as $s^{src}_t \in R^{hw\times fhw}$, where $f=2$ is the number of frames used as key and value. It captures both the structure of a single frame and the temporal correspondence with the warped frames.

% For zero-shot appearance editing, we set the warped index as the middle frame $\Call{w}{i} = \text{Round}[\frac{n}{2}]$.
% For shape editing, using the previous frame and the first one ~\cite{tuneavideo}
% $\Call{w}{i} = [1, i-1]$ is better.

% \noindent\textbf{Editing Method Summary.}
Overall, the proposed method produces a new editing method for zero-shot real-world video editing. We replace the attention maps in the denoising steps with their corresponding maps during the inversion steps. After that, we utilize cross-attention maps as masks to prevent semantic leaks. Finally, we reform the self-attention of UNet to spatial-temporal attention for better temporal consistency among different temporal frames. We have included a formal algorithm in the supplementary materials for reference purposes.

% 3. what solution we choose: mask the self attention in low resolution layer 8X8, 16X16 using layout of cross attention

% 4. what is our implementation
% write the math formation

% \circled{1}

% \subsection{Video Style \& Attribution Editing}

% CAR: 
% 
% 4. torun: under sky, desert sky with replacement

% boat:
% 1. to run: iced water 2. 

% rabit
% pizza;

% cartoon

% suppliment 24 frames

% suppliment 24 frames

\begin{figure*}[h]
\centering

\newcommand{\imwidth}{1.0\textwidth}
% \setcellgapes{0.5em}
% \makegapedcells
\vspace{-1em}
\begin{tabular}{@{}c@{}}
  % Prompt Driving Video (top) and Result (bottom) \\

% {a silver jeep driving down a snow-covered road in the countryside} &
% {a snow-covered road} &
\parbox{\imwidth}{\includegraphics[width=\imwidth]{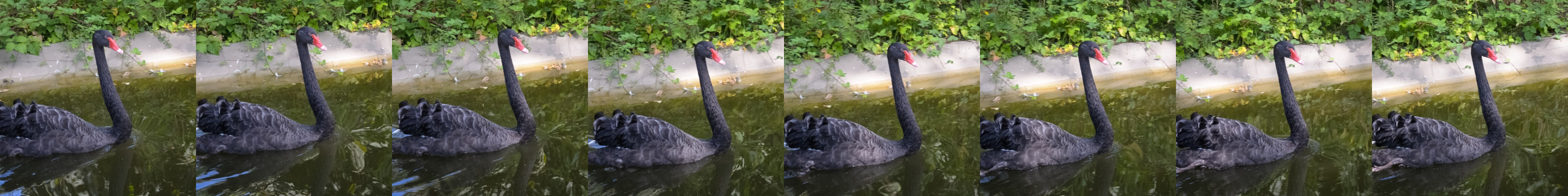}}
\\
{Source Prompt: 
% \texttt{A black swan with a red beak swimming in a river near a wall and bushes.}
{A black swan with a red beak swimming in a river near a wall and bushes.}
}
% \parbox{\imwidth}{\includegraphics[width=\imwidth, ]{figs/imgs/main_results/cartoon_swan.png}}
% \\
% {Zero-shot style editing on pre-trained image diffusion model~\cite{stable-diffusion}: \texttt{A cartoon photo of} ... 
% }
\\
\parbox{\imwidth}{\includegraphics[width=\imwidth, ]{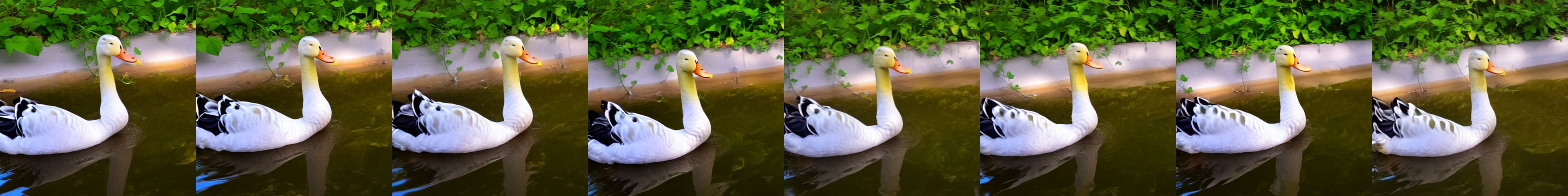}}
\\
{
% Zero-shot object shape editing on pre-trained video diffusion model~\cite{tuneavideo}: 
% \texttt{black swan} $\xrightarrow{}$ \texttt{white duck}.
{black swan} $\xrightarrow{}$ \textcolor{red}{white duck}.
}
\\
\parbox{\imwidth}{\includegraphics[width=\imwidth, ]{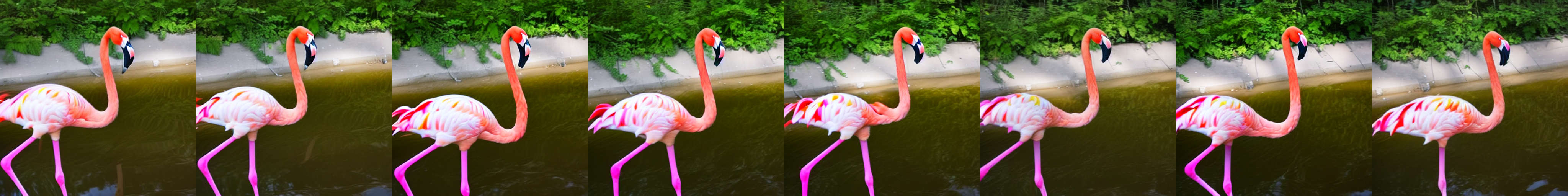}}
\\
{
% Zero-shot object shape editing on pre-trained video diffusion model~\cite{tuneavideo}: 
% \texttt{black swan} $\xrightarrow{}$ \texttt{pink flamingo}.
{black swan} $\xrightarrow{}$ \textcolor{red}{pink flamingo}.}
\end{tabular}
% \vspace{-1em}
  \caption{\textbf{Zero-shot object shape editing on pre-trained video diffusion model~\cite{tuneavideo}:} Our framework can directly edit the shape of the object in videos driven by text prompts using a trained video diffusion model~\cite{tuneavideo} 
  % \chenyang{add red bbox on the background to denote the motion detail preservation; remove cartoon if no space}
  }
  % \vspace{1em}
  \label{fig:exp_swan_shape_edit}
\end{figure*}%

\begin{figure*}[h]
\centering

\newcommand{\imwidth}{1.0\textwidth}
% \setcellgapes{0.5em}
% \makegapedcells
\vspace{-1em}
\begin{tabular}{@{}c@{}}
% Prompt Driving Video (top) and Result (bottom) 
% \\
\parbox{\imwidth}{\includegraphics[width=\imwidth, ]{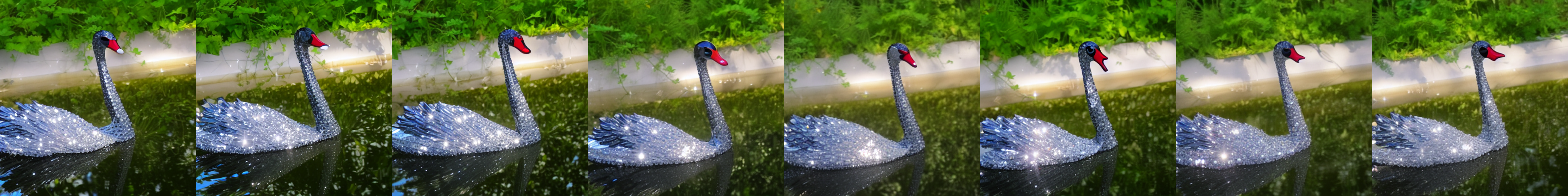}}
\\
% { Source Prompt from Fig~\ref{fig:exp_swan_shape_edit}: \texttt{black} $\xrightarrow{}$ \texttt{Swarovski crystal} 
{ Source Prompt from Fig~\ref{fig:exp_swan_shape_edit}: black $\xrightarrow{}$ \textcolor{red}{Swarovski crystal}} 
% {\includegraphics[width=\imwidth]{figs/imgs/main_results/rabit_pokemon.png}}
% \\
% {A rabbit is eating a watermelon $\rightarrow$ \textcolor{red}{Pokemon cartoon of} a rabbit ...}
\\
\parbox{\imwidth}{\includegraphics[width=\imwidth]{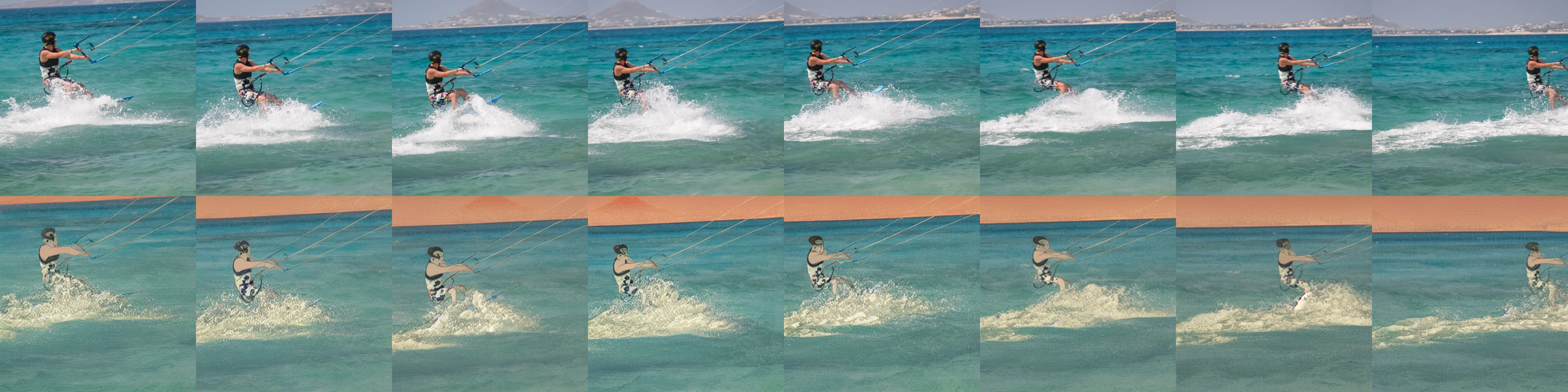}}
\\
{A man with round helmet surfing on a white wave $\rightarrow$ 
\textcolor{red}{The Ukiyo-e style painting of } a man ...}
\\
\parbox{\imwidth}{\includegraphics[width=\imwidth]{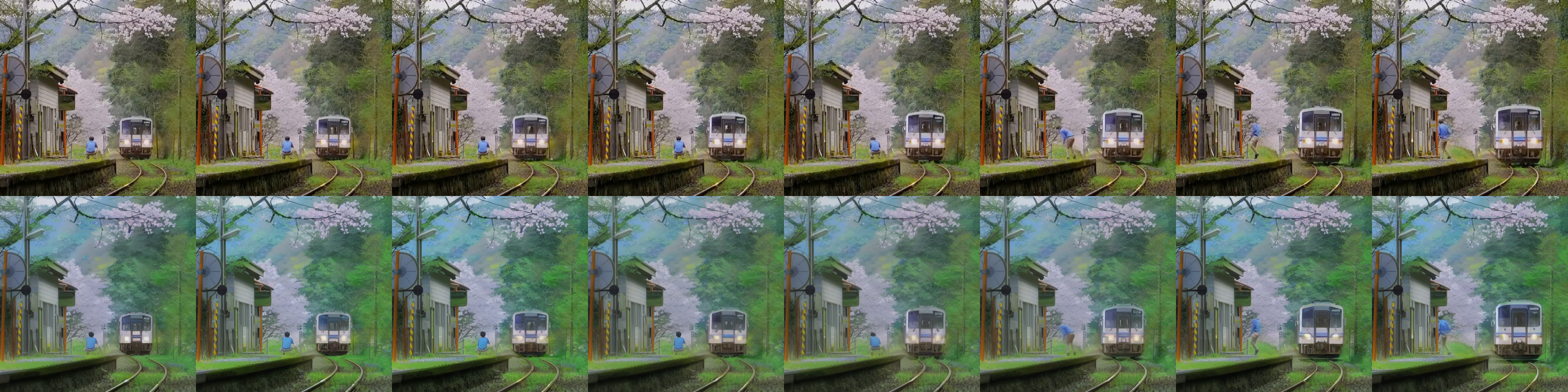}}
\\
{A train traveling down tracks next to a forest and a man on the side of the track $\rightarrow$ 
% {A train traveling down tracks next to a forest and a man on the side of the track $\rightarrow$ 
..., \textcolor{red}{Makoto Shinkai style}}
\\
\end{tabular}
% \vspace{-1em}
  \caption{\textbf{Zero-shot attribute and style editing results using Stable Diffusion~\cite{stable-diffusion}.} Our framework supports abstract attribute and style editing like `Swarovski crystal', `Ukiyo-e', and `Makoto Shinkai'. Best viewed with zoom-in.
  }
  \label{fig:exp_attribute_style_edit}
\end{figure*}%
\subsection{Shape-Aware Video Editing}
\label{sec:shape-aware-editing}
\begin{figure*}[t!]
    \centering
    \includegraphics[width=1.\textwidth, 
    % height=0.4\textwidth
    ]{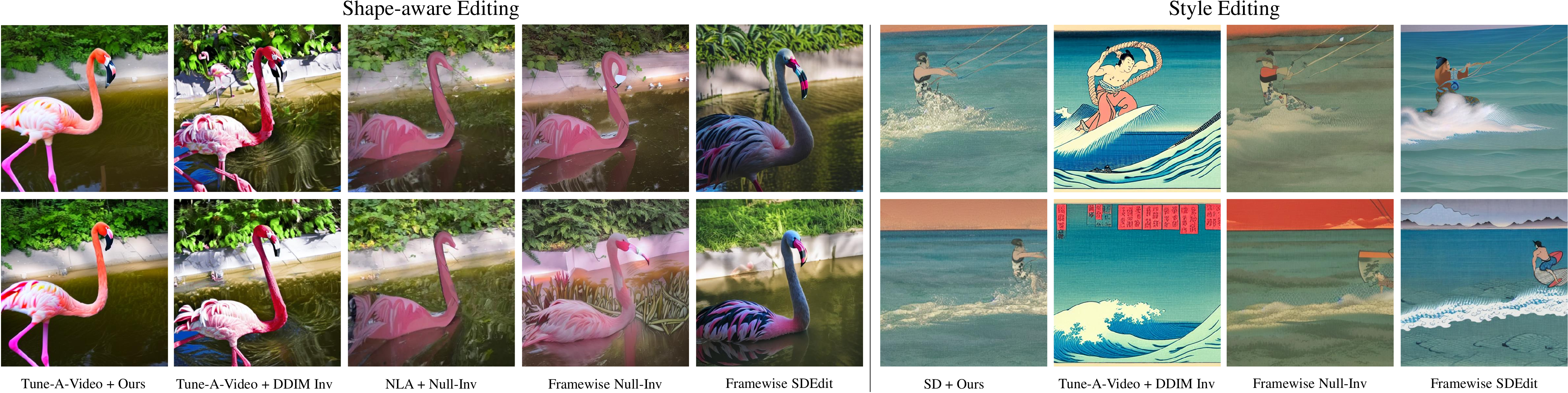} % compare_w_ours.pdf
    \vspace{-1em}
    \caption{\textbf{Qualitative comparison of our methods with other baselines.} Inputs are in Fig.~\ref{fig:exp_swan_shape_edit} and Fig~\ref{fig:exp_attribute_style_edit}. Our results have the best temporal consistency, image fidelity, and editing quality. Best viewed with zoom-in.}
    % \vspace{-1em}
    \label{fig:baseline}
\end{figure*}

Different from appearance editing, reforming the shape of a specific object in the video is much more challenging. To this end, a pretrained video diffusion model is needed. Since there is no publicly-available generic video diffusion model, we perform the editing on the one-shot video diffusion model~\cite{tuneavideo} instead. In this case, we compare our editing method with simple DDIM inversion~\cite{ddim}, where our method also achieves better performance in terms of editing ability, motion consistency, and temporal consistency. It might be because it is hard for an inflated model to overfit the exact motion of the input video. While in our method, the motion and structure are represented by high-quality spatial-temporal attention maps $s^{src}_t \in R^{hw\times fhw}$ during inversion, which is further fused with the attention maps during editing. More details can be founded in Fig.~\ref{fig:baseline} and the supp. video.

\section{Experiments}

\label{sec:exp}
% CAR: 
% Teaser 1. porsche 2. watercolor 
% 3. snow-covered 4. torun: under sky

% boat:
% 1. to run: iced water 2. 

% rabit
% pizza; pokemon latter; 

% sur
% 0226_surf_50_style_ukiyo_640_230227-013245
% cartoon, swarovski
% suppliment 24 frames
% suppliment 24 frames

\subsection{Implementation Details}
For zero-shot style and attribute editing, we directly use the trained stable diffusion v1.4~\cite{stable-diffusion} as the base model, we fuse the attentions in the interval of  $t\in[0.2\times T, T]$ of the DDIM step with total timestep $T=50$.
For shape editing, we utilize the pretrained model of the specific video~\cite{tuneavideo} at 100 iterations and fuse the attention at DDIM timestep $ t \in[0.5\times T, T]$, giving more freedom for new shape generation. Following previous works~\cite{gen1, text2live}, we use videos from DAVIS~\cite{davis} and other in-the-wild videos to evaluate our approach. The source prompt of the video is generated via the image caption model~\cite{BLIP}. Finally, we design the target prompt for each video by replacing or adding several words.
\vspace{-0.5em}
\subsection{Applications}
\noindent\textbf{Local attribute and global style editing.} Using pretrained text-to-image diffusion model~\cite{stable-diffusion}, our framework supports zero-shot local attribute and global style editing, as shown in Fig.~\ref{fig:exp_attribute_style_edit} and third row in Fig.\ref{fig:teaser}. 
In the first row, the texture and color of the feather are modified
by the target prompt \texttt{Swarovski crystal} and kept consistent across frames. In the second and third rows, our framework applies abstract style (\texttt{Ukiyo-e} and \texttt{Makoto Shinkai}). The image structure and temporal motion can be well preserved since we fuse both the spatial-temporal self-attention and cross-attention during the inversion and editing stage.

\noindent\textbf{Shape-aware editing.} Fig.~\ref{fig:exp_swan_shape_edit} and the second row in Fig.\ref{fig:teaser} present the result of difficult object shape editing, with a pretrained video model~\cite{tuneavideo}. This task is challenging because a naive full-resolution fusion of the spatial-temporal self-attention maps results in inaccurate shape results and wrong temporal motion, as shown in the ablation (Fig.\ref{fig:ablation_masked_attention}). Thanks to the proposed Attention Blending, we combine the motion of generated shape from the editing target and inverted attention from the input video. Results of \texttt{posche}, \texttt{duck} and \texttt{flamingo} show that we generate new content with poses and positions similar to input videos.
% \label{fig:exp_swan_shape_edit}

\noindent\textbf{Zero-shot image editing.} In addition, our framework can serve as a zero-shot image editing method such as local attribute editing (Fig.~\ref{fig:attention comparison}) and object shape editing (Fig.~\ref{fig: attention mixing}) by considering an image as a video with a single frame.
% local attribute editing
% object shape editing
% overall style editing
% zero-shot image enhancement (optinal)
% zero-shot object removal (optinal)
We provide more results in our supplementary material.
% zero-shot image enhancement (optinal)

% zero-shot object removal (optinal)
% \newpage
% \input{figs/main_result_shape_editing}
% page 5
% \newpage

\begin{table}[t]

\centering
\resizebox{\linewidth}{!}{
\begin{tabular}{@{}l@{\hspace{2mm}}c@{\hspace{2mm}}*{3}{c@{\hspace{2mm}}}r@{}}
% \begin{tabular}{@{}l@{\hspace{5mm}}c@{\hspace{1.5mm}}c@{\hspace{5mm}}c@{\hspace{4mm}}c@{\hspace{4mm}}c@{\hspace{3mm}}c@{\hspace{1mm}}c@{\hspace{1mm}}c@{\hspace{1mm}}c@{\hspace{1mm}}}

\toprule

% \small{Method & \multicolumn{2}{c}{Neural Metrics} & \multicolumn{3}{c}{User study}} \\
% Method & \multicolumn{2}{c}{Neural Metrics} & \multicolumn{2}{c}{User study} \\
Method & \multicolumn{2}{c}{CLIP Metrics$\uparrow$}& \multicolumn{3}{c}{User Study$\downarrow$} \\
% &
% \multicolumn{6}{c}{Reconstructed HR PSNR$\uparrow$} \\
\cmidrule(l{1mm}r{1mm}){2-3} 
\cmidrule(l{1mm}r{1mm}){4-6} 
% \small{Inversion\& Editing & Clip-score & FVD & Editing  & Image-Quality & Temp-Consis } 
Inversion \& Editing & Tem-Con & Fram-Acc & Edit  & Image & Temp \\

% \cmidrule(l{1mm}r{1mm}){7-13}
\midrule
% NeuralAtlases~\cite{layeraltas} \& NullInv~\cite{null} \\
\small{Framewise Null \& p2p\cite{null,p2p}}  & 0.852 & \textbf{0.958} & 3.55 & 4.11 & 4.38 \\
\small{Framewise SDEit\cite{sdedit}} & 0.910 & 0.819 & 3.69 & 3.28 & 3.62 \\
\small{NLA, Null \& p2p\cite{layeraltas,null,p2p}} & 0.949 & 0.600 & 3.17 & 3.02 & 2.60\\
\small{Tune-A-Video \& DDIM\cite{tuneavideo,ddim}} & 0.958  & 0.750 & 2.78 & 2.80 &  2.70  \\
% \small{Framewise Null\cite{null}} \& p2p\cite{p2p}  & 0.852 & \textbf{0.958} & 3.55 & 4.11 & 4.38 \\
% \small{Framewise SDEit\cite{sdedit}} & 0.910 & 0.819 & 3.69 & 3.28 & 3.62 \\
% \small{NLA\cite{layeraltas}, Null\cite{null} \& p2p\cite{p2p}} & 0.949 & 0.600 & 3.17 & 3.02 & 2.60\\
% \small{Tune-A-Video\cite{tuneavideo} \& DDIM\cite{ddim}} & 0.958  & 0.750 & 2.78 & 2.80 &  2.70  \\
\midrule

Ours & \textbf{0.965} & 0.903 & \textbf{1.82} & \textbf{1.79} & \textbf{1.69} \\

\bottomrule
\end{tabular}
}
% \vspace{-1em}
\caption{\textbf{Quantitative evaluation against baselines.} In our user study, the results of our method are preferred over those from baselines. For CLIP-Score, we achieve the best temporal consistency and comparable framewise editing accuracy against an optimization-based image editing method~\cite{null}.}
\label{table:Quantitative_baseline}
% \vspace{-1em}
\end{table}

\subsection{Baseline Comparisons}
Since there are no available zero-shot video editing methods based on diffusion models, we build the following four state-of-the-art baselines for comparison. (1)~Tune-A-Video~\cite{tuneavideo} overfits an inflated diffusion model on a single video to generate similar content. 
% During shape editing, we find our method can get better editing results using the earlier checkpoints.
% is the similar to use. 
(2) The Neural Layered Atlas~\cite{layeraltas}~(NLA) based method is combined with keyframe-editing via state-of-the-art image editing methods~\cite{null,p2p}.
(3) Frame-wise Null-text optimization~\cite{null} and then edit by prompt2prompt~\cite{p2p}.
(4) Frame-wise zero-shot editing using SDEdit~\cite{sdedit}.
For attention-based editing~(2,3,4), we use the same timesteps fusion parameters as ours.

We conduct the quantitative evaluation using the trained CLIP~\cite{CLIP1} model as previous methods~\cite{gen1,tuneavideo,pix2pix-zero}. Specially, we show the \textbf{`Tem-Con'}~\cite{gen1} to measure the temporal consistency in frames by computing the cosine similarity between all pairs of consecutive frames. \textbf{`Frame-Acc'}~\cite{pix2pix-zero,CLIP1,CLIPScore} is the frame-wise editing accuracy, which is the percentage of frames where the edited image has a higher CLIP similarity to the target prompt than the source prompt. 
In addition, three user studies metrics~(denoted as \textbf{`Edit'}, \textbf{`Image'}, and \textbf{`Temp'}) are conducted to measure the editing quality, overall frame-wise image fidelity, and temporal consistency of the video, respectively. We ask 20 subjects to rank different methods with 9 sets of comparisons in each study. 
From Tab.~\ref{table:Quantitative_baseline}, the proposed zero-shot method achieves the best temporal consistency against baselines and shows a comparable frame-wise editing accuracy as the pre-frame optimization method~\cite{null}. As for the user studies, the average ranking of our method earns user preferences the best in three aspects. 

\begin{figure}[t]
    \centering
    \includegraphics
    [width=0.47\textwidth]
    {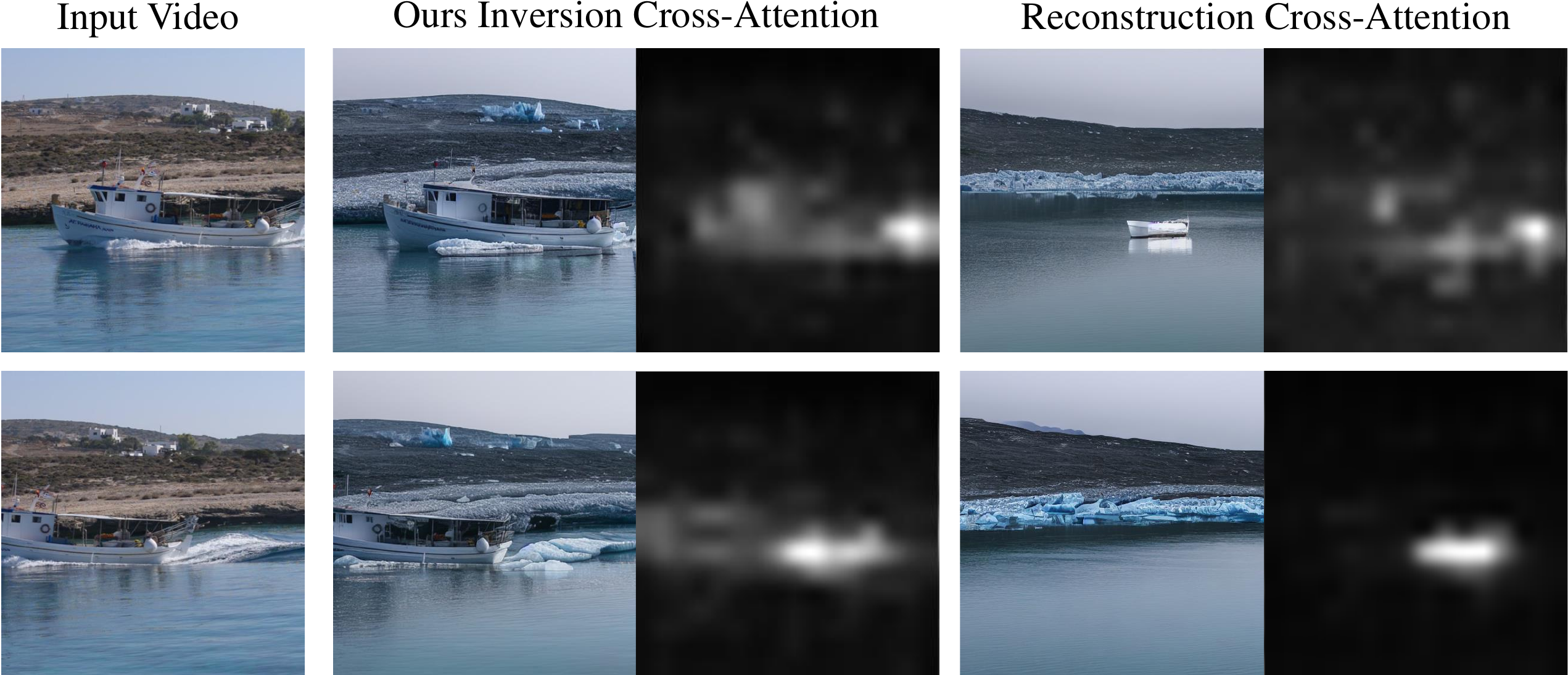}
    \caption{ \textbf{Inversion attention compared with reconstruction attention using prompt `deserted shore $\xrightarrow{}$ \textcolor{red}{`glacier shore'}.} The attention maps obtained from the reconstruction stage fail to detect the boat's position, and can not provide suitable motion guidance for zero-shot video editing. 
    % \xiaodong{the purple has some semantic leak in the road, might be replaced to another example}
    } 
    % \vspace{-2em}
    \label{fig:ablation_inversion_attn}
\end{figure}

To provide a qualitative comparison, Fig.\ref{fig:baseline} provides the results of our method and other baselines at two different frames.
% The results of framewise editing methods , since these baselines ignore the temporal relationship.
% Framewise SDEdit~\cite{sdedit} is an efficient zero-shot method
The editing result of framewise SDEdit~\cite{sdedit} can not be localized and varies a lot among different frames. Frame-wise Null inversion achieves local editing at the cost of 500-iterations optimization for each frame but is still temporally inconsistent.
% Although it has the best quantitative `Frame-Acc', its temporal consistency is much worse than other methods in both CLP metrics and user study, because it optimizes each frame individually.
NLA-based~\cite{layeraltas} method 
% ranks second in a temporal-consistency user study and 
preserves the exact pixels in the atlas. However, it struggles to perform editing that involves new shapes or 3D structures. In addition, it takes hours to optimize the neural atlas for each input video.
% Recently, Tune-a-video~\cite{tuneavideo} proposes to tune an inflated stable diffusion model on a single video. 
While Tune-A-Video~\cite{tuneavideo} with DDIM~\cite{ddim} ranks second in editing quality and image fidelity of Tab.~\ref{table:Quantitative_baseline}, we observe that it has difficulty in reproducing the exact motion and spatial position as input video (right side of Fig.\ref{fig:baseline}). Besides, the background has annoying artifacts. Different from the above baselines, our method preserves the motion by fusion the attention during inversion and editing. Thus, our results outperform others by a large margin in our user study and frame consistency measured by CLIP.
% with comparable frame-wise editing accuracy as the image editing method~\cite{null}.

\subsection{Ablation Studies}
Although we have proved the effectiveness of the proposed strategies in Fig.~\ref{fig: attention mixing} and Fig.~\ref{fig:attention comparison} using toy image examples, here, we ablate these designs in the video.

\noindent\textbf{Attention during inversion.} In the right column of Fig.~\ref{fig:ablation_inversion_attn}, we use the attention map during reconstruction instead of inversion for zero-shot background editing. The visualized cross-attention map of the word \texttt{`boat'} in the first and last frame can not capture the correct position and structure of the boat,
% The motion of the object is also inconsistent between the input video and generated video,
which may be caused by the poor temporal modeling capacity of the image diffusion model and the accumulation of errors in DDIM inversion. In contrast, we propose using attention during inversion as the middle column, which provides stable guidance of semantic layout in the original video. We observe this huge difference in attention maps between inversion and reconstruction exists in most videos.
% during our experiments.
% \twocolumn[{
% \maketitle
% \begin{center}
%     \captionsetup{type=figure}
%     \vspace{-2em}
%     \includegraphics[draft,width=1.\textwidth, height=0.3\textwidth]{figs/images/teaser.pdf}
%     \vspace{-2em}
%         \captionof{figure}{Qualitative comparison of our methods with other baselines.}
% \end{center}
% }]
\begin{figure}[t]
    \centering
    \includegraphics
    [width=0.47\textwidth]
    {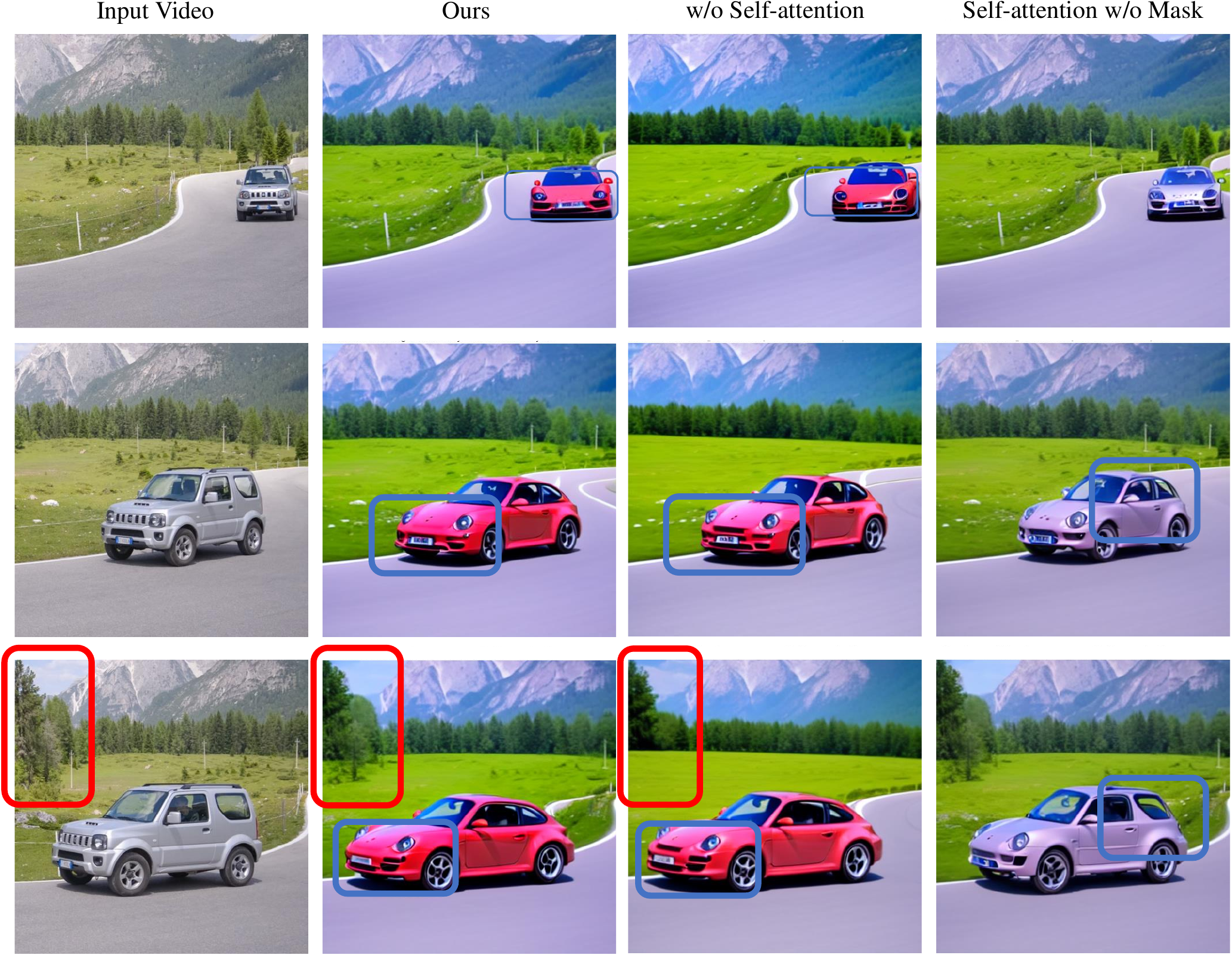}
    \caption{\textbf{Ablation study of blended self-attention.} Without self-attention fusion, the generated video can not preserve the details of input videos (e.g., fence, trees, and car identity).
    % and the identity of the car changes in different timesteps. 
    If we replace full self-attention without a spatial mask, the structure of the original jeep misleads the generation of the Porsche car.}
    % \vspace{-2em}
    \label{fig:ablation_masked_attention}
\end{figure}

\noindent\textbf{Attention Blending Block} is studied in 
Fig.~\ref{fig:ablation_masked_attention}, where we remove all self-attention fusion or fuse all self-attention without a spatial mask. The third column shows that removing all self-attention maps brings a loss of fine details ( \eg, fences, poles, and trees in the background) and inconsistency of car identity over time. In contrast, if we fuse full-resolution self-attention as in the previous work~\cite{p2p}, the shape editing ability of the framework can be severely degraded so that the geometry of generated car resembles the input video, especially in the last few frames. Therefore, we propose to blend the self-attention maps with a mask obtained from cross-attention to preserve unedited details and ensure temporal consistency while editing the object shape.

% newpage

% newpage

% \newpage
% page 7
% \newpage
% page 8

% \newpage
% \subsection{Limitation}
% % \xiaodong{ablation study of the temporal-conv, failed case when changing bird to the flying dinosaur.}
% % \chenyang{Move this section to supp? It does not present our contribution. }
% % page 8.5
% % \newpage

% While our method achieves impressive results in video editing, it still has some limitations. During shape editing, since the motion is leaned by the one-shot video diffusion model~\cite{tuneavideo}, it is difficult to generate totally new motion~(\eg, `swimming' $\xrightarrow{}$ `fly' ) or very different shape~(\eg, `swan' $\xrightarrow{}$ 'pterosaur'). We believe a stronger video diffusion model might solve these problems.
% (b) Our editing capacity is bounded by the performance of pretrained-model, which bring obstacles to generate diverse movie styles as gen1~\cite{gen1} (\eg, claymation)

\section{Conclusion}
In this paper, we propose a new text-driven video editing framework \texttt{FateZero} that performs temporal consistent zero-shot editing of attribute, style, and shape. 
We make the first attempt to study and utilize the cross-attention and spatial-temporal self-attention during DDIM inversion, which provides fine-grained motion and structure guidance at each denoising step.
A new Attention Blending Block is further proposed to enhance the shape editing performance of our framework.
Our framework benefits \textbf{video} editing using widely existing \textbf{image} diffusion models, which we believe will contribute to a lot of new video applications.

\noindent \textbf{Limitation \& Future Work.}
While our method achieves impressive results,
% in video editing, 
it still has some limitations. During shape editing, since the motion is produced by the one-shot video diffusion model~\cite{tuneavideo}, it is difficult to generate totally new motion~(\eg,`swim'$\xrightarrow{}$`fly' ) or very different shape~(\eg,`swan' $\xrightarrow{}$`pterosaur'). We will test our method on the generic pretrained video diffusion model for better editing abilities.

\noindent \textbf{Acknowledgement}
% This work was carried out during an internship at Tencent AI Lab.
This project is supported by the National Key R\&D Program of China under grant number 2022ZD0161501.
The authors would like to express sincere gratitude to Tencent AI Lab for providing the necessary computation resources and a conducive environment for research. 
 Additionally, the authors extend their appreciation to Xilin Zhang for reviewing and revising the writing, and to all friends at Tencent and HKUST who participated in the user study.
% We leave the application of our techniques to other pretrained image diffusion models~\cite{controlnet} as future work.
\label{sec:conclusion}

{\small
\bibliographystyle{ieee_fullname}
\bibliography{11_references}

\begin{thebibliography}{10}\itemsep=-1pt

\bibitem{civitai_website}
https://civitai.com, 2020.

\bibitem{blended_latent}
Omri Avrahami, Ohad Fried, and Dani Lischinski.
\newblock Blended latent diffusion.
\newblock {\em arXiv preprint arXiv:2206.02779}, 2022.

\bibitem{blended}
Omri Avrahami, Dani Lischinski, and Ohad Fried.
\newblock Blended diffusion for text-driven editing of natural images.
\newblock In {\em Proceedings of the IEEE/CVF Conference on Computer Vision and
  Pattern Recognition}, pages 18208--18218, 2022.

\bibitem{text2live}
Omer Bar-Tal, Dolev Ofri-Amar, Rafail Fridman, Yoni Kasten, and Tali Dekel.
\newblock Text2live: Text-driven layered image and video editing.
\newblock In {\em European Conference on Computer Vision}, pages 707--723.
  Springer, 2022.

\bibitem{BTSSPP15}
Nicolas Bonneel, James Tompkin, Kalyan Sunkavalli, Deqing Sun, Sylvain Paris,
  and Hanspeter Pfister.
\newblock Blind video temporal consistency.
\newblock {\em ACM Transactions on Graphics (Proceedings of SIGGRAPH Asia
  2015)}, 34(6), 2015.

\bibitem{diffedit}
Guillaume Couairon, Jakob Verbeek, Holger Schwenk, and Matthieu Cord.
\newblock Diffedit: Diffusion-based semantic image editing with mask guidance.
\newblock {\em arXiv preprint arXiv:2210.11427}, 2022.

\bibitem{DiffusionBeatGANs}
Prafulla Dhariwal and Alex Nichol.
\newblock Diffusion models beat gans on image synthesis.
\newblock {\em Neural Information Processing Systems}, 2021.

\bibitem{ding2021cogview}
Ming Ding, Zhuoyi Yang, Wenyi Hong, Wendi Zheng, Chang Zhou, Da Yin, Junyang
  Lin, Xu Zou, Zhou Shao, Hongxia Yang, et~al.
\newblock Cogview: Mastering text-to-image generation via transformers.
\newblock {\em Advances in Neural Information Processing Systems},
  34:19822--19835, 2021.

\bibitem{gen1}
Patrick Esser, Johnathan Chiu, Parmida Atighehchian, Jonathan Granskog, and
  Anastasis Germanidis.
\newblock Structure and content-guided video synthesis with diffusion models.
\newblock {\em arXiv preprint arXiv:2302.03011}, 2023.

\bibitem{taming}
Patrick Esser, Robin Rombach, and Bjorn Ommer.
\newblock Taming transformers for high-resolution image synthesis.
\newblock In {\em Proceedings of the IEEE/CVF conference on computer vision and
  pattern recognition}, pages 12873--12883, 2021.

\bibitem{stylit}
Jakub Fi{\v{s}}er, Ond{\v{r}}ej Jamri{\v{s}}ka, Michal Luk{\'a}{\v{c}}, Eli
  Shechtman, Paul Asente, Jingwan Lu, and Daniel S{\`y}kora.
\newblock Stylit: illumination-guided example-based stylization of 3d
  renderings.
\newblock {\em ACM Transactions on Graphics (TOG)}, 35(4):1--11, 2016.

\bibitem{fivser2017example}
Jakub Fi{\v{s}}er, Ond{\v{r}}ej Jamri{\v{s}}ka, David Simons, Eli Shechtman,
  Jingwan Lu, Paul Asente, Michal Luk{\'a}{\v{c}}, and Daniel S{\`y}kora.
\newblock Example-based synthesis of stylized facial animations.
\newblock {\em ACM Transactions on Graphics (TOG)}, 36(4):1--11, 2017.

\bibitem{gatys2016image}
Leon~A Gatys, Alexander~S Ecker, and Matthias Bethge.
\newblock Image style transfer using convolutional neural networks.
\newblock In {\em Proceedings of the IEEE conference on computer vision and
  pattern recognition}, pages 2414--2423, 2016.

\bibitem{gan}
Ian Goodfellow, Jean Pouget-Abadie, Mehdi Mirza, Bing Xu, David Warde-Farley,
  Sherjil Ozair, Aaron Courville, and Yoshua Bengio.
\newblock Generative adversarial networks.
\newblock {\em Communications of the ACM}, 63(11):139--144, 2020.

\bibitem{lvdm}
Yingqing He, Tianyu Yang, Yong Zhang, Ying Shan, and Qifeng Chen.
\newblock Latent video diffusion models for high-fidelity video generation with
  arbitrary lengths.
\newblock {\em arXiv preprint arXiv:2211.13221}, 2022.

\bibitem{p2p}
Amir Hertz, Ron Mokady, Jay Tenenbaum, Kfir Aberman, Yael Pritch, and Daniel
  Cohen-Or.
\newblock Prompt-to-prompt image editing with cross attention control.
\newblock {\em arXiv preprint arXiv:2208.01626}, 2022.

\bibitem{CLIPScore}
Jack Hessel, Ari Holtzman, Maxwell Forbes, Ronan~Le Bras, and Yejin Choi.
\newblock Clipscore: A reference-free evaluation metric for image captioning.
\newblock {\em Empirical Methods in Natural Language Processing}, 2021.

\bibitem{imagen-video}
Jonathan Ho, William Chan, Chitwan Saharia, Jay Whang, Ruiqi Gao, Alexey
  Gritsenko, Diederik~P Kingma, Ben Poole, Mohammad Norouzi, David~J Fleet,
  et~al.
\newblock Imagen video: High definition video generation with diffusion models.
\newblock {\em arXiv preprint arXiv:2210.02303}, 2022.

\bibitem{ddpm}
Jonathan Ho, Ajay Jain, and Pieter Abbeel.
\newblock Denoising diffusion probabilistic models.
\newblock {\em Advances in Neural Information Processing Systems},
  33:6840--6851, 2020.

\bibitem{vdm}
Jonathan Ho, Tim Salimans, Alexey Gritsenko, William Chan, Mohammad Norouzi,
  and David~J Fleet.
\newblock Video diffusion models.
\newblock {\em arXiv:2204.03458}, 2022.

\bibitem{ebsynth}
Ond\v{r}ej Jamri\v{s}ka, \v{S}\'{a}rka Sochorov\'{a}, Ond\v{r}ej Texler, Michal
  Luk\'{a}\v{c}, Jakub Fi\v{s}er, Jingwan Lu, Eli Shechtman, and Daniel
  S\'{y}kora.
\newblock Stylizing video by example.
\newblock {\em ACM Trans. Graph.}, 38(4), jul 2019.

\bibitem{johnson2016perceptual}
Justin Johnson, Alexandre Alahi, and Li Fei-Fei.
\newblock Perceptual losses for real-time style transfer and super-resolution.
\newblock In {\em Computer Vision--ECCV 2016: 14th European Conference,
  Amsterdam, The Netherlands, October 11-14, 2016, Proceedings, Part II 14},
  pages 694--711. Springer, 2016.

\bibitem{stylegan}
Tero Karras, Samuli Laine, and Timo Aila.
\newblock A style-based generator architecture for generative adversarial
  networks.
\newblock In {\em Proceedings of the IEEE/CVF conference on computer vision and
  pattern recognition}, pages 4401--4410, 2019.

\bibitem{layeraltas}
Yoni Kasten, Dolev Ofri, Oliver Wang, and Tali Dekel.
\newblock Layered neural atlases for consistent video editing.
\newblock {\em ACM Transactions on Graphics (TOG)}, 40(6):1--12, 2021.

\bibitem{imagic}
Bahjat Kawar, Shiran Zada, Oran Lang, Omer Tov, Huiwen Chang, Tali Dekel, Inbar
  Mosseri, and Michal Irani.
\newblock Imagic: Text-based real image editing with diffusion models.
\newblock {\em arXiv preprint arXiv:2210.09276}, 2022.

\bibitem{vae}
Diederik~P Kingma and Max Welling.
\newblock Auto-encoding variational bayes.
\newblock {\em arXiv preprint arXiv:1312.6114}, 2013.

\bibitem{lai2018learning}
Wei-Sheng Lai, Jia-Bin Huang, Oliver Wang, Eli Shechtman, Ersin Yumer, and
  Ming-Hsuan Yang.
\newblock Learning blind video temporal consistency.
\newblock In {\em Proceedings of the European conference on computer vision
  (ECCV)}, pages 170--185, 2018.

\bibitem{shape-aware-editing}
Yao-Chih Lee, Ji-Ze Genevieve~Jang Jang, Yi-Ting Chen, Elizabeth Qiu, and
  Jia-Bin Huang.
\newblock Shape-aware text-driven layered video editing demo.
\newblock {\em arXiv preprint arXiv:2301.13173}, 2023.

\bibitem{dvp}
Chenyang Lei, Yazhou Xing, and Qifeng Chen.
\newblock Blind video temporal consistency via deep video prior.
\newblock In {\em Advances in Neural Information Processing Systems}, 2020.

\bibitem{lei2022deep}
Chenyang Lei, Yazhou Xing, Hao Ouyang, and Qifeng Chen.
\newblock Deep video prior for video consistency and propagation.
\newblock {\em IEEE Transactions on Pattern Analysis and Machine Intelligence},
  45(1):356--371, 2022.

\bibitem{BLIP}
Junnan Li, Dongxu Li, Caiming Xiong, and Steven Hoi.
\newblock Blip: Bootstrapping language-image pre-training for unified
  vision-language understanding and generation.
\newblock 2023.

\bibitem{sdedit}
Chenlin Meng, Yang Song, Jiaming Song, Jiajun Wu, Jun-Yan Zhu, and Stefano
  Ermon.
\newblock Sdedit: Image synthesis and editing with stochastic differential
  equations.
\newblock {\em arXiv preprint arXiv:2108.01073}, 2021.

\bibitem{CLIP1}
Alexander~H. Miller, Will Feng, Dhruva Tirumala, Adam Fisch, Augustus Odena,
  Vivek Ramavajjala, Joel~Z. Leibo, Kelvin~Guu andJesse Engel, Jack Clark,
  Maruan~H. Ali, Nazneen Rajani, Iain~J. Dunning, Jacob Andreas, Chris Dyer,
  Dario Amodei, Jakob Uszkoreit, Douwe Pieksma, Tom Brown, and Ilya Sutskever.
\newblock Clip: Learning to solve visual tasks by unsupervised learning of
  language representations.
\newblock In {\em International Conference on Machine Learning}, 2020.

\bibitem{null}
Ron Mokady, Amir Hertz, Kfir Aberman, Yael Pritch, and Daniel Cohen-Or.
\newblock Null-text inversion for editing real images using guided diffusion
  models.
\newblock {\em arXiv preprint arXiv:2211.09794}, 2022.

\bibitem{dreamix}
Eyal Molad, Eliahu Horwitz, Dani Valevski, Alex~Rav Acha, Yossi Matias, Yael
  Pritch, Yaniv Leviathan, and Yedid Hoshen.
\newblock Dreamix: Video diffusion models are general video editors.
\newblock {\em arXiv preprint arXiv:2302.01329}, 2023.

\bibitem{t2i-adaptor}
Chong Mou, Xintao Wang, Liangbin Xie, Jian Zhang, Zhongang Qi, Ying Shan, and
  Xiaohu Qie.
\newblock T2i-adapter: Learning adapters to dig out more controllable ability
  for text-to-image diffusion models.
\newblock {\em arXiv preprint arXiv:2302.08453}, 2023.

\bibitem{pix2pix-zero}
Gaurav Parmar, Krishna~Kumar Singh, Richard Zhang, Yijun Li, Jingwan Lu, and
  Jun-Yan Zhu.
\newblock Zero-shot image-to-image translation.
\newblock {\em arXiv preprint arXiv:2302.03027}, 2023.

\bibitem{davis}
Jordi Pont-Tuset, Federico Perazzi, Sergi Caelles, Pablo Arbel{\'a}ez,
  Alexander Sorkine-Hornung, and Luc~Van Gool.
\newblock The 2017 davis challenge on video object segmentation.
\newblock {\em arXiv: Computer Vision and Pattern Recognition}, 2017.

\bibitem{dalle2}
Aditya Ramesh, Prafulla Dhariwal, Alex Nichol, Casey Chu, and Mark Chen.
\newblock Hierarchical text-conditional image generation with clip latents.
\newblock {\em arXiv preprint arXiv:2204.06125}, 2022.

\bibitem{ramesh2021zero}
Aditya Ramesh, Mikhail Pavlov, Gabriel Goh, Scott Gray, Chelsea Voss, Alec
  Radford, Mark Chen, and Ilya Sutskever.
\newblock Zero-shot text-to-image generation.
\newblock In {\em International Conference on Machine Learning}, pages
  8821--8831. PMLR, 2021.

\bibitem{stable-diffusion}
Robin Rombach, Andreas Blattmann, Dominik Lorenz, Patrick Esser, and Björn
  Ommer.
\newblock High-resolution image synthesis with latent diffusion models, 2021.

\bibitem{unet}
Olaf Ronneberger, Philipp Fischer, and Thomas Brox.
\newblock U-net: Convolutional networks for biomedical image segmentation.
\newblock In {\em Medical Image Computing and Computer-Assisted
  Intervention--MICCAI 2015: 18th International Conference, Munich, Germany,
  October 5-9, 2015, Proceedings, Part III 18}, pages 234--241. Springer, 2015.

\bibitem{imagen}
Chitwan Saharia, William Chan, Saurabh Saxena, Lala Li, Jay Whang, Emily
  Denton, Seyed Kamyar~Seyed Ghasemipour, Burcu~Karagol Ayan, S~Sara Mahdavi,
  Rapha~Gontijo Lopes, et~al.
\newblock Photorealistic text-to-image diffusion models with deep language
  understanding.
\newblock {\em arXiv preprint arXiv:2205.11487}, 2022.

\bibitem{make-a-video}
Uriel Singer, Adam Polyak, Thomas Hayes, Xi Yin, Jie An, Songyang Zhang, Qiyuan
  Hu, Harry Yang, Oron Ashual, Oran Gafni, et~al.
\newblock Make-a-video: Text-to-video generation without text-video data.
\newblock {\em arXiv preprint arXiv:2209.14792}, 2022.

\bibitem{ddim}
Jiaming Song, Chenlin Meng, and Stefano Ermon.
\newblock Denoising diffusion implicit models.
\newblock {\em arXiv preprint arXiv:2010.02502}, 2020.

\bibitem{ncsn}
Yang Song and Stefano Ermon.
\newblock Generative modeling by estimating gradients of the data distribution.
\newblock In {\em Advances in Neural Information Processing Systems}, pages
  11895--11907, 2019.

\bibitem{pnp}
Narek Tumanyan, Michal Geyer, Shai Bagon, and Tali Dekel.
\newblock Plug-and-play diffusion features for text-driven image-to-image
  translation.
\newblock {\em arXiv preprint arXiv:2211.12572}, 2022.

\bibitem{vqvae}
Aaron Van Den~Oord, Oriol Vinyals, et~al.
\newblock Neural discrete representation learning.
\newblock {\em Advances in neural information processing systems}, 30, 2017.

\bibitem{attention_is_all_you_need}
Ashish Vaswani, Noam Shazeer, Niki Parmar, Jakob Uszkoreit, Llion Jones,
  Aidan~N Gomez, {\L}ukasz Kaiser, and Illia Polosukhin.
\newblock Attention is all you need.
\newblock {\em Advances in neural information processing systems}, 30, 2017.

\bibitem{Phenaki}
Ruben Villegas, Mohammad Babaeizadeh, Pieter-Jan Kindermans, Hernan Moraldo,
  Han Zhang, Mohammad~Taghi Saffar, Santiago Castro, Julius Kunze, and Dumitru
  Erhan.
\newblock Phenaki: Variable length video generation from open domain textual
  descriptions.
\newblock In {\em International Conference on Learning Representations}, 2023.

\bibitem{tuneavideo}
Jay~Zhangjie Wu, Yixiao Ge, Xintao Wang, Stan~Weixian Lei, Yuchao Gu, Wynne
  Hsu, Ying Shan, Xiaohu Qie, and Mike~Zheng Shou.
\newblock Tune-a-video: One-shot tuning of image diffusion models for
  text-to-video generation.
\newblock {\em arXiv preprint arXiv:2212.11565}, 2022.

\bibitem{Vtoonify}
Shuai Yang, Liming Jiang, Ziwei Liu, and Chen~Change Loy.
\newblock Vtoonify: Controllable high-resolution portrait video style transfer.
\newblock {\em ACM Transactions on Graphics (TOG)}, 41(6):1--15, 2022.

\bibitem{controlnet}
Lvmin Zhang and Maneesh Agrawala.
\newblock Adding conditional control to text-to-image diffusion models, 2023.

\bibitem{zhang2018perceptual}
Richard Zhang, Phillip Isola, Alexei~A Efros, Eli Shechtman, and Oliver Wang.
\newblock The unreasonable effectiveness of deep features as a perceptual
  metric.
\newblock In {\em CVPR}, 2018.

\bibitem{magic-video}
Daquan Zhou, Weimin Wang, Hanshu Yan, Weiwei Lv, Yizhe Zhu, and Jiashi Feng.
\newblock Magicvideo: Efficient video generation with latent diffusion models.
\newblock {\em arXiv preprint arXiv:2211.11018}, 2022.

\end{thebibliography}
}

\ifarxiv \clearpage \appendix
\label{sec:appendix}

% \section*{Summary}
% \noindent This supplementary material is organized as follows.
% \begin{itemize}
%   \renewcommand{\labelitemi}{\textbullet}

%  % \item Section~\ref{Sec:Details of Bit Rate Loss} presents our design of bit rate loss. \chenyang{Maybe introduce more details about auxiliary loss and entropy model if we have time}

% \item Section~\ref{sec:Implementation Details} provides the details and implementations of our architecture.
% \item Section~\ref{sec:demo_video} introduces our supplementary demo video.
% % \item Section~\ref{sec:baseline comparisons} shows more comparison with previous works. \chenyang{temporal profile}
% % \item Section~\ref{sec:Additional application} presents other side applications of our methods (\eg, long video editing, video enhancement, object removal \& inpainting, single image editing, local editing with blending mask, video control-net editing)
% % \item The `supplement.mp4' in the package gives a video visualization of our method and results.
% \item Section~\ref{sec:limitation} discusses the potential improvement of our work.
% % /home/chenyangqi/disk1/fast_rescaling/Video-Enhancement-Playground/results/1013_f24_4x_2e2_ccm24_no_crop

% \end{itemize}

\section{Implementation Details}
\label{sec:Implementation Details}
\noindent \textbf{Pseudo algorithm code} Our full algorithm is shown in Algorithm~\ref{alg:real_image_editing} and Algorithm~\ref{alg:attention_fusion}. Algorithm~\ref{alg:real_image_editing} presents the overall framework of our inversion and editing, as visualized in the left of Fig. \textcolor{red}{1} in the main paper.
Algorithm~\ref{alg:attention_fusion} shows that the cross-attention is fused based on a mask of the edited words, and the self-attention is blended using a binary mask from thresholding the cross-attention (the right of Fig.~\textcolor{red}{1} in the main paper).

\noindent \textbf{Hyperparameters Tuning}. There are mainly three hyperparameters in our proposed designs:
 \\
 - ${t}_s \in [1, T]$: Last timestep of the self-attention blending. Smaller ${t}_s$ fuses more self-attention from inversion to preserve structure and motion.
 \\
 - ${t}_c \in [1, T]$: Last timestep of the cross attention fusion. Smaller ${t}_c$ fuses more cross attention from inversion to preserve the spatial semantic layout.
 \\
 - $\tau\in [0, 1]$: Threshold for the blending mask used in shape editing. Smaller $\tau$ uses more self-attention map from editing to improve shape editing results.

 In \textbf{style} and \textbf{attribute} editing, we set ${t}_s=0.2T$, ${t}_c=0.3T$, $\tau=1.0$ to preserve most structure and motion in the source video.
 In \textbf{shape} editing, we set ${t}_s=0.5T$, ${t}_c=0.5T$, $\tau=0.3$ to give more freedom in new motion and 3D shape generation.

\section{Demo Video}
\label{sec:demo_video}
% \chenyang{More video results about style, attribute, shape}
\noindent we provide a detailed demo video to show:

\noindent\textbf{Video Results} on style, local attribute, and shape editing to validate the effectiveness of the proposed method.

\noindent\textbf{Method Animation} to provide a better understanding of the proposed method.

\noindent\textbf{Baseline Comparisons} with previous methods in video.

\noindent\textbf{More Promising Applications} We have shown the effectiveness of the proposed method in the main paper for style, attribution, and shape editing. In the demo video, we also show some potential applications of the proposed method, including (1) object removal by removing the word of the target object in the source prompt and mask the self-attention of the corresponding area using its cross attention, (2) video enhancement by adding the specific prompt~(\eg, `high-quality', `8K') in the target editing prompt.
~\label{sec:Additional application}

\newpage
\section{Limitation and Future Work}

\algnewcommand{\ElseIIf}[1]{\algorithmicelse\ #1}
\begin{algorithm}[t]
\caption{ \texttt{FateZero} Algorithm}
\label{alg:real_image_editing}
\begin{algorithmic}

\State \textbf{Input:} 
\\ 
- $z_0$: \text{Latent code from source video}
\\
- $p_{src}$: \text{Source text prompt for input video}
\\
- $p_{edit}$: \text{Target text prompt for edition}
\\
\State {\bf Hyperparameters:} 
\\
 - ${t}_c$: Last timestep of the cross attention fusion
 \\
 - ${t}_s$: Last timestep of the self attention blending
 \\
 - $\tau$: Threshold for blending mask

\\
 \State {\bf Output:} 
 \\
 - $\hat{z}_0$: \text{Final edited latent code}

\\ \\
 $\triangleright$ DDIM for inversion latents and attention maps

% \xiaodong{check $p_{edit}$ or $p_{src}$}

\For{$t = 1,2,...,T$}
    \State $\epsilon_t, c_{t}^{\text{src}}, s_{t}^{\text{src}} \gets  \epsilon_\theta(z_t, t, p_{src})$
    % \State $z_{t+1}=\Call{Inversion\_step}{z_t, \hat{\epsilon}, t}$
    \State $z_{t} = \sqrt{\alpha_{t}} \; \frac{z_{t-1} - \sqrt{1-\alpha_{t-1}}\epsilon_t}{\sqrt{\alpha_{t-1}}}+ \sqrt{1-\alpha_{t}}\epsilon_t$
\EndFor

\\
\State $\triangleright$ Denoising the inverted latents with attention fusion

\For{$t = T, (T-1),...,1$}
    % \State $\_\_, M_{t}^{\text{edit}} \gets \epsilon_\theta(z_t, t, p_{\text{edit}})$
    
    \State $\text{Edited\_index} = (p_{src} \text{ !=\ \ } \ p_{edit})$
    % \State $M_{\text{cross}} = (p_{src} \text{ !=\ \ } \ p_{edit})$
    \State $\triangleright$ Cross-attention mask is from the edited index~\cite{p2p}
    \State $M_{\text{cross}}[\text{Edited\_index}] = 1$
    \State $\triangleright$ Self-attention blending mask is from cross-attention.
    \State $M_{\text{self}} =  (c_{t}^{\text{src}}[\text{Edited\_index}] > \tau)$
    \State $\hat{\epsilon_t} \gets \Call{Att-Fusion}{\varepsilon_\theta, z_t, t, p_{\text{edit}}, M_{\text{edit}}, M_{\text{self}}, c_{t}^{\text{src}}, s_{t}^{\text{src}}}$
    % \State $z_{t-1}=\Call{Denoising\_step}{z_t, \hat{\epsilon}, t}$
    \State $z_{t-1} = \sqrt{\alpha_{t-1}} \; \frac{z_t - \sqrt{1-\alpha_t}\hat{\epsilon_t}}{\sqrt{\alpha_t}}+ \sqrt{1-\alpha_{t-1}}\hat{\epsilon_t}$
\EndFor

\State $\triangleright$ Fuse the inversion and editing attention of all $B$ blocks.
\State $\triangleright$ We only show the operation of attention and omit the feed-forward, residual convolution layer for simplicity.
% \Function{Att-Fusion}{$\theta, z_t, t, p_{\text{edit}}, c_{t}^{\text{src}}, s_{t}^{\text{src}}$}
\Function{Att-Fusion}{$\varepsilon_\theta, z_t, t, p_{\text{edit}}, M_{\text{cross}}, M_{\text{self}}, c_{t}^{\text{src}}, s_{t}^{\text{src}}$}
\For{$i = 1...B$}
    \State $s_{t}^{\text{edit}} = \text{Softmax}(W_i^Q(z_{t})W_i^K(z_{t})/\sqrt{d_i} )$
    \State $s_{t}^{\text{fused}} = \Call{Self-Blending}{s_{t}^{\text{edit}}, s_{t}^{\text{src}}, M_{\text{self}}, c_{t}^{\text{src}}, t}$
    \State $z_{t} \ \ \ \ = W_i^V(z_{t})\cdot s_{t}^{\text{fused}}$
    \State $c_{t}^{\text{edit}} = \text{Softmax}(W_i^Q(z_{t})W_i^K(p_{edit})/\sqrt{d_i} )$
    \State $c_{t}^{\text{fused}} = \Call{Cross-Fusion}{c_{t}^{\text{edit}}, c_{t}^{\text{src}}, M_{\text{edit}}, t}$
    \State $z_{t} \ \ \ \ = W_i^V(p_{\text{edit}})\cdot c_{t}^{\text{fused}}$
        
\EndFor
\State \Return $z_t$
\EndFunction

\end{algorithmic}
\end{algorithm}

\begin{algorithm}[t]
\caption{Attention Fusion and Blending Algorithm}
\label{alg:attention_fusion}
\begin{algorithmic}

\\
\State $\triangleright$ Cross-attention fusion using the difference mask between source and editing prompt following prompt-to-prompt.
\Function{Cross-Fusion}{$c_{t}^{\text{edit}}, c_{t}^{\text{src}}, M_{\text{edit}}, t$}
\If{$t > t_c $} 
\State \Return $ M_{\text{cross}} \cdot c_{t}^{\text{edit}}  + (1-M_{\text{cross}}) \cdot c_{t}^{\text{src}}$
% \\
\Else \State \Return $c_{t}^{\text{edit}}$
% \Else  1
% \ElseIf{$3$} $4$
\EndIf
\EndFunction
\\
\State $\triangleright$ Self-attention blending with cross attention.
\Function{Slef-Blending}{$s_{t}^{\text{edit}}, s_{t}^{\text{src}}, c_{t}^{\text{src}}, M_{\text{self}}, t$}
\If{$t > t_s $} 
\State \Return $ M_{\text{self}} \cdot s_{t}^{\text{edit}}  + (1-M_{\text{self}}) \cdot s_{t}^{\text{src}}$
% \\
\Else \State \Return $s_{t}^{\text{edit}}$
% \Else  1
% \ElseIf{$3$} $4$
\EndIf
\EndFunction

\end{algorithmic}
\end{algorithm}

Our zero-shot editing is not good at new concept composition or generation of very different shapes. For example, the result of editing `black swan' to `yellow pterosaur' in Fig~\ref{fig:limitation} is unsatisfactory. This problem may be alleviated using a stronger video diffusion model, which we leave to future work.
\label{sec:limitation}
% \xiaodong{ablation study of the temporal-conv, failed case when changing bird to the flying dinosaur.}
\begin{figure}[t]
\centering

\newcommand{\imwidth}{0.45\textwidth}
% \setcellgapes{0.5em}
% \makegapedcells
% \vspace{-1em}
\begin{tabular}{@{}c@{}}
  % Prompt Driving Video (top) and Result (bottom) \\
\parbox{\imwidth}{\includegraphics[width=\imwidth, ]{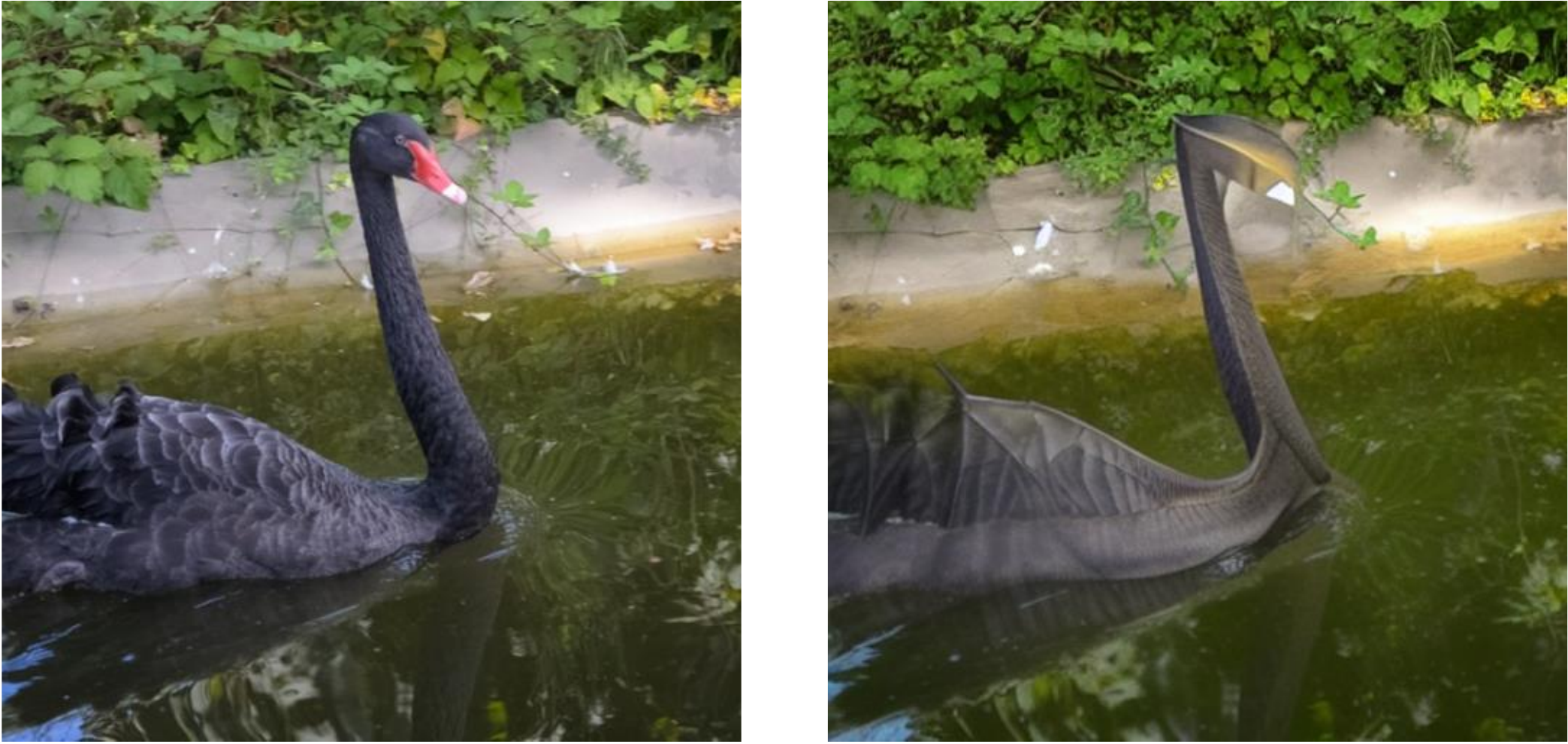}}
\\
{
% Zero-shot object shape editing on pre-trained video diffusion model~\cite{tuneavideo}: 
% \texttt{black swan} $\xrightarrow{}$ \texttt{pink flamingo}.
{black swan} $\xrightarrow{}$ \textcolor{red}{ yellow pterosaur}.
}
\end{tabular}
% \vspace{-1em}
\caption{limitation of our zero-shot editing.
  % \chenyang{add red bbox on the background to denote the motion detail preservation; remove cartoon if no space}
}
  % \vspace{1em}
  \label{fig:limitation}
\end{figure}%

 \fi

\end{document}